\documentclass[12pt]{article}
\usepackage{amssymb,amsmath,amsthm,epsfig}
\usepackage{cite}
\usepackage{eepic}
\usepackage{epic}
\usepackage{graphicx}
\usepackage{color}
\usepackage{ifpdf}
\usepackage[title]{appendix}
\usepackage{amsthm}

\usepackage{dsfont}
\usepackage{graphicx} 
\usepackage{subfigure}

\usepackage{algpseudocode}
\usepackage{algorithmicx, algorithm}

\usepackage{amsmath}
\usepackage{bm}
\usepackage{multirow}

\usepackage{caption}

\usepackage{tikz}
\usetikzlibrary{shapes.geometric,arrows}
\usepackage{makecell}

\usepackage{graphicx}
\usepackage{epstopdf}
\usepackage[colorlinks=true,
linkcolor=blue,
anchorcolor=blue,
citecolor=blue,
urlcolor=blue,]{hyperref}
\usepackage{cleveref}
\usepackage{notoccite}

\theoremstyle{plain}
\newtheorem{lem}{Lemma}[section]

\newtheorem{prop}[lem]{Proposition}

\theoremstyle{definition}
\newtheorem{defn}{Definition}[section]

\theoremstyle{remark}
\newtheorem{rem}{Remark}[section]

\begin{document}
	
	\title{\large\bf {{Reinforcement learning based data assimilation for unknown state model}}}
	\author{Ziyi Wang\thanks{School of Mathematical Sciences,  Tongji University, Shanghai 200092, China. ({\tt 2111156@tongji.edu.cn}).}
		\and
		Lijian Jiang\thanks{School of Mathematical Sciences,  Tongji University, Shanghai 200092, China. ({\tt  ljjiang@tongji.edu.cn}).}
	}
\date{}
\maketitle
\begin{abstract}
	Data assimilation (DA) has increasingly emerged as a critical tool for state estimation across a wide range of applications. It is  significantly challenging when the governing equations of the underlying dynamics are unknown. To this end, various machine learning approaches have been employed to construct a surrogate state transition model in a supervised learning framework, which  relies on pre-computed training datasets. However, it is often infeasible to obtain noise-free ground-truth state sequences  in practice. To address this challenge, we propose a novel method that integrates reinforcement learning with ensemble-based Bayesian filtering methods, enabling the learning of surrogate state transition model for unknown dynamics directly from noisy observations, without using  true state trajectories. Specifically, we treat the process for computing maximum likelihood estimation of surrogate model parameters as a sequential decision-making problem, which can be formulated as a discrete-time Markov decision process (MDP). Under this formulation, learning the surrogate transition model is equivalent to finding an optimal policy of the MDP, which can be effectively addressed using reinforcement learning techniques. Once the model is trained offline, state estimation can be performed in the online stage using filtering methods based on the learned dynamics. The proposed framework accommodates a wide range of observation scenarios, including nonlinear and partially observed measurement models. A few numerical examples  demonstrate that the proposed method achieves superior accuracy and robustness in high-dimensional settings.

\end{abstract}\smallskip

\smallskip

{\bf keywords:}
data assimilation, reinforcement learning, data-driven modeling
\section{Introduction}
Estimating the state of a dynamical system from noisy observations is of significance  in data assimilation (DA) \cite{DA-introduction,DA-application}, which has  broad applications in areas such as signal processing \cite{DA-signal}, weather forecasting \cite{DA-weather}, and oil recovery \cite{DA-oil}. Traditional DA methods assume that the state transitional model of the dynamical system are known and perform state estimation by integrating observation data with model information. For example, Kalman filter (KF)\cite{KF} yields an optimal state estimate for linear state-space models with additive Gaussian noise through closed-form solutions. In nonlinear settings where the KF becomes inadequate, ensemble-based Bayesian filtering methods, such as ensemble Kalman filter (EnKF) \cite{EnKF,EnKF-1} and particle filter (PF) \cite{PF-1}, employ an ensemble of particles to approximate the probability distribution of the target state. These model-based approaches rely on the availability of an explicit and accurate state transition model of the system dynamics. However, deriving governing equations for complex dynamical systems remains a significant challenge \cite{modeling-1,modeling-2,modeling-3}, particularly when the underlying physical laws are limited or unknown.

\begin{figure}[htbp]
	\centering
	\includegraphics[scale=0.13]{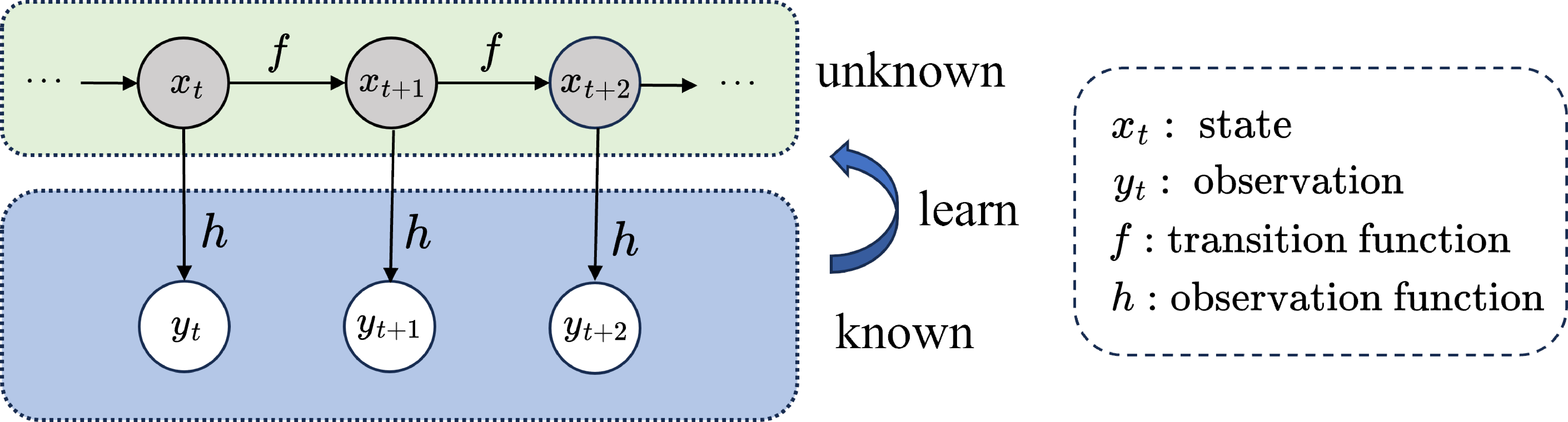}
	\caption{Problem setting of the proposed methods.}
	\label{problem-setting}
\end{figure}

Recently, machine learning has been increasingly employed to construct data-driven state transition models for DA. When some state trajectories are available, surrogate models can be trained in a supervised learning framework \cite{RKN,Takens,FBF}, after which DA can be performed on the learned dynamics. However, in many real-world applications, acquiring noise-free ground-truth state sequences is often infeasible \cite{obs-1,obs-2}. In contrast, observations governed by linear or nonlinear observation functions are typically more accessible (see Figure \ref{problem-setting}). In such cases, it becomes necessary to learn the surrogate state transition model directly from noisy observations, particularly when the observation function is nonlinear.

Unsupervised learning-based methods have demonstrated significant potential for addressing this issue. Inspired by the effectiveness of ensemble-based DA methods in addressing nonlinear problems, one could  incorporate these techniques into the expectation-maximization (EM) algorithm \cite{EM,EM-EnKS-1,EM-EnKS-2}. In this framework, the expectation step (E-step) approximates the posterior distribution using ensemble-based approaches, such as ensemble Kalman smoother \cite{EnKS} or particle smoother \cite{PS}. The maximization step then updates a neural network (NN) surrogate model based on the state estimates obtained from the E-step. The accuracy of the E-step approximation is critical to the overall performance of these EM-based methods. However, achieving an accurate approximation of the E-step may be challenging, particularly in high-dimensional and strongly nonlinear settings. To address this limitation, another approach \cite{VI} employs variational inference to jointly optimize a parameterized posterior distribution and model parameters. Major challenges in this framework include designing effective posterior parameterizations and selecting robust initialization strategies to ensure training stability.

Reinforcement learning (RL) \cite{RL} is a paradigm of machine learning in which an agent learns to make decisions by interacting with an environment to maximize cumulative reward. In contrast to supervised learning, which relies on a pre-computed labeled dataset for training, RL is inherently suited for sequential decision-making through trial-and-error interactions. Recent advances in deep reinforcement learning \cite{DRL,DRL-1} have further extended its applicability to high-dimensional state spaces, enabling a wide range of applications such as game playing \cite{RL-game,RL-game-2}, robotics \cite{RL-robtics}, and autonomous driving\cite{RL-auto}. Notably, several studies \cite{RL-DA,RL-DA-1} have investigated the integration of RL with DA. However, most existing studies focus on scenarios where the governing equations of the system are known, thereby restricting their applicability to model-free settings where system dynamics are unknown or difficult to characterize.

In this paper, we propose a novel framework for DA in scenarios where the underlying dynamical model is  unknown, by combining RL with ensemble-based Bayesian filtering methods such as EnKF and PF. In the offline stage, we learn a surrogate state transition model directly from noisy observations, without using  the true state trajectories. The key idea of this framework is to reformulate the process for computing maximum likelihood estimation (MLE) of model parameters as a sequential decision-making problem, which can be modeled as a discrete-time Markov decision process (MDP). This MDP consists of four key components: (1) the state, which comprises the posterior particles of filtering distribution at the current time step and the observation at the next time step; (2) the action, which includes the prior particles of forecast distribution and the observation at the next time step; (3) the state transition, which is determined by the analysis step of Bayesian filtering; (4) the reward, representing the estimation of log-marginal-likelihood for one time step. The objective of this MDP is to find an optimal policy that maximizes the expected cumulative reward, which serves as an estimator of the log-marginal-likelihood of the observations. Under this formulation, learning the surrogate transition model is equivalent to solving the corresponding MDP, which can be addressed using some RL techniques. It is worth noting that the proposed framework is compatible with a wide range of observation scenarios, including nonlinear and partially observed measurement models. Once the surrogate model is learned offline, state estimation can be performed in the online stage using filtering methods based on the learned dynamics. Specially, we employ the proximal policy optimization (PPO) \cite{PPO} algorithm to solve the MDP, and propose  two specific implementations of the proposed framework:  PPO based EnKF (PPO-EnKF) and PPO based PF (PPO-PF), respectively.

The paper is organized as follows. In Section \ref{Preliminaries}, we present the problem setting and provide a brief overview of Bayesian filtering and RL. In Section \ref{MDP-MLE}, we present how to reformulate the MLE problem as an MDP. In Section \ref{RL-DA}, we employ the PPO algorithm to solve the MDP and propose the PPO-EnKF and PPO-PF methods for DA without the state transition model. In Section \ref{results}, we demonstrate the performance of the proposed methods by some numerical examples. Finally, concluding remarks are provided in Section \ref{conclusions}.

\section{Preliminaries}
\label{Preliminaries}
\subsection{Bayesian filtering}
Consider the following discrete-time state-space model (SSM):
\begin{equation}
	\begin{cases}
		x_{t+1}= f (x_{t}) +\xi_t,\quad \xi_t \sim \mathcal{N}(0,Q),\\
		y_{t+1}=h(x_{t+1})+\eta_{t+1},\quad \eta_{t+1} \sim \mathcal{N}(0,R),\\
	\end{cases}
	\label{ssm}
\end{equation}
where $x_{t+1} \in \mathbb{R}^m$ is the model state evolving according to the state transition function $f$, and $y_{t+1} \in \mathbb{R}^n$ is the observation generated by the observation function $h$. The process noise $\xi_t$ and observation noise $\eta_{t+1}$ are assumed to be i.i.d. Gaussian random variables with covariances $Q$ and $R$, respectively. For a given SSM, Bayesian filtering provides a recursive framework to estimate the filtering distribution $p(x_{t+1}|y_{1:t+1})$. This process proceeds from time  $t$ to $t+1$ via two key steps: the forecast step and the analysis step.
\begin{itemize}
	\item \textbf{Forecast step.} Estimate the forecast distribution $p(x_{t+1}|y_{1:t})$ using the Chapman-Kolmogorov equation:
	\begin{equation}
		p(x_{t+1}|y_{1:t}) = \int p(x_{t+1}|x_{t}) p(x_{t}|y_{1:t}) \mathrm{d}x_{t}.
		\label{da-fore}
	\end{equation}
	
	\item \textbf{Analysis step.} Derive the filtering distribution $p(x_{t+1}| y_{1:t+1})$ via Bayes’ formula:
	\begin{equation}
		p(x_{t+1}|y_{1:t+1}) = \frac{p(y_{t+1}|x_{t+1})p(x_{t+1}|y_{1:t})}{p(y_{t+1}|y_{1:t})}
		\propto p(y_{t+1}|x_{t+1})p(x_{t+1}|y_{1:t}).
		\label{da-analysis}
	\end{equation}
\end{itemize}
By iteratively applying (\ref{da-fore}-\ref{da-analysis}), the filtering distribution $p(x_{t+1}|y_{1:t+1})$ can be recursively updated over time. However, for nonlinear SSMs, obtaining analytical solutions to these equations is generally intractable. To address this challenge, some ensemble-based methods have been developed to approximate the filtering distribution using a finite set of particles, among which a popular choice is the EnKF.

The EnKF \cite{EnKF} is a sequential Monte Carlo method that approximates the filtering distribution using an ensemble of equally weighted particles. Let $\hat{x}_t^{1:N} = \{\hat{x}_t^{(i)}\}_{i=1}^N$ and $x_t^{1:N} = \{x_t^{(i)}\}_{i=1}^N$ represent the forecast and analysis ensembles, respectively. The main steps of the EnKF is described as follows:
\begin{itemize}
	\item \textbf{Forecast step.} Generate the prior particle $\hat{x}_{t+1}^{(i)}$ through the state transition model defined in (\ref{ssm}):
	\begin{equation}
		\begin{aligned}
			\hat{x}_{t+1}^{(i)} &= f(x_{t}^{(i)}) + \xi_t^{(i)},\quad \xi_t^{(i)}\sim \mathcal{N}(0,Q).\\
		\end{aligned}
		\label{EnKF-fore}
	\end{equation}
	\item \textbf{Analysis step.} Apply a Kalman-type update to calculate the posterior particle $x_{t+1}^{(i)}$:
	\begin{equation}
		\begin{aligned}
			\hat{m}_{t+1} &= \frac{1}{N} \sum_{i=1}^{N} \hat{x}_{t+1}^{(i)},\quad \hat{n}_{t+1} = \frac{1}{N} \sum_{i=1}^{N} h(\hat{x}_{t+1}^{(i)}),\\
			X_{t+1} &= \left[\hat{x}_{t+1}^{(1)}-\hat{m}_{t+1}, \cdots, \hat{x}_{t+1}^{(N)}-\hat{m}_{t+1}\right],\\
			Y_{t+1} &= \left[h(\hat{x}_{t+1}^{(1)})-\hat{n}_{t+1},\cdots,h(\hat{x}_{t+1}^{(N)})-\hat{n}_{t+1}\right],\\
			K_{t+1} &= \frac{X_{t+1} Y_{t+1}^{\top}}{N-1} \left( \frac{Y_{t+1} Y_{t+1}^{\top}}{N-1} + R\right)^{-1},\\
			x_{t+1}^{(i)} &= \hat{x}_{t+1}^{(i)} + K_{t+1}(y_{t+1} + \eta_{t+1}^{(i)} - h(\hat{x}_{t+1}^{(i)})), \quad \eta_{t+1}^{(i)}\sim \mathcal{N}(0,R).
		\end{aligned}
		\label{EnKF-analysis}
	\end{equation}
\end{itemize}
In the analysis step, the observation $y_{t+1}$ is perturbed by noise term $\eta_{t+1}^{(i)} \sim \mathcal{N}(0,R)$. This perturbation ensures that, under the assumption of a linear and Gaussian SSM, the empirical mean and covariance of $x_{t+1}^{1:N}$ converge to the mean and covariance of the true filtering distribution as $N \rightarrow \infty$ \cite{EnKF}.

\subsection{Problem setting}
Classical filtering methods require an explicit state transition model for state estimation. However, in many practical scenarios, the underlying state dynamics are unknown or difficult to model explicitly. To address this challenge, we propose a DA framework that does not rely on prior knowledge of the state transition model. In this setting, only noisy observations $y_{1:T}$ are available, obtained through the observation model:
$$
y_{t+1}=h(x_{t+1})+\eta_{t+1},\quad \eta_{t+1} \sim \mathcal{N}(0,R),
$$
where the observation function $h$ may be linear or nonlinear.

In the offline stage, we aim to learn a surrogate state transition model for the unknown dynamics:
\begin{equation}
	x_{t+1}= F^{\mathrm{NN}}_{\alpha} (x_t) +\xi_t,\quad \xi_t \sim \mathcal{N}(0,Q_{\beta}),
	\label{transition-model}
\end{equation}
where $F^{\mathrm{NN}}_{\alpha}$ denotes an NN parameterized by $\alpha$, and the process noise covariance $Q_{\beta}$ is parameterized by $\beta$. We formulate the process for learning the model parameters $\theta = \{\alpha, \beta\}$ as a sequential decision-making problem, which can be solved within RL framework. Once the surrogate model (\ref{transition-model}) is trained, it can be used for DA during the online phase.

\subsection{Reinforcement learning}
Before describing the proposed methods, we provide a brief review of RL. RL tasks can be modeled as discrete-time MDPs \cite{RL}, denoted as $\mathcal{M} = \{\mathcal{S}, \mathcal{A}, P, r\}$, where $\mathcal{S}$ and $\mathcal{A}$ represent the state and action spaces, respectively. The state transition probability $P:\mathcal{S} \times \mathcal{A} \times \mathcal{S} \rightarrow \mathbb{R}$ defines the distribution of the next state $s_{t+1}$ given the current state $s_t$ and action $a_t$. The function $r: \mathcal{S} \times \mathcal{A} \rightarrow \mathbb{R}$ specifies the reward for a given state-action pair.

At each time step, the agent receives a state $s_t$ and selects an action $a_t$ according to its policy $\pi(a_t|s_t)$, which maps states to a probability distribution over the action space. After executing the action and interacting with the environment, the agent transitions to the next state $s_{t+1}$ and receives a scalar reward $r(s_t,a_t)$ generated by the environment. In a finite-horizon MDP, the interaction between the agent and the environment over an episode yields a trajectory  $\{(s_{t},a_{t})\}_{t=0}^{T'-1}$, along with the corresponding sequence of rewards $\{r(s_t,a_t)\}_{t=0}^{T'-1}$, where $T'$ represents the terminal time step. The discounted cumulative reward of the trajectory is defined as
$$
R = \sum_{t=0}^{T'-1} \gamma^{t} r(s_t,a_t),
$$
where $\gamma \in [0,1]$ is the discount rate that balances the importance of immediate and future rewards. Since the trajectory is influenced by the stochasticity of the policy and environment, $R$ is a random variable. Consequently, the performance of a policy $\pi$ is evaluated by its expected discounted cumulative reward:
$$
\begin{aligned}
	J(\pi) = \mathbb{E}_{s_0,a_0,\cdots,s_{T'-1},a_{T'-1}} \left[R\right],
\end{aligned}
$$
and the objective of RL is to find an optimal policy $\pi^*$ that maximizes $J(\pi)$.

Policy-based methods form a fundamental class of RL algorithms that directly optimize the policy, in contrast to value-based methods, which estimate value functions to derive a policy. In policy-based approaches, the policy is parameterized by a set of learnable parameters $\omega$, yielding a parametric policy $\pi_{\omega}(a_t|s_t)$. The corresponding optimization objective becomes:
\begin{equation}
	\omega^{*} = \arg \max_{\omega} J(\pi_{\omega}).
	\label{RL-objective}
\end{equation}
In this paper, we employ the PPO algorithm, a widely used policy-based method known for its stability and effectiveness in high-dimensional continuous control problems.

\section{Formulating the MLE problem as an MDP}
\label{MDP-MLE}
In this section, we reformulate the MLE of $\theta$ as an MDP. Subsection \ref{data-log} describes the estimation of the observation log-marginal-likelihood function $\log p(y_{1:T};\theta)$ using filtering methods. A detailed formulation of the MDP is presented in Subsection \ref{MDP}.

\subsection{Estimation of the data log-marginal-likelihood}
\label{data-log}
Given the observation trajectory $y_{1:T}$, the observation log-marginal-likelihood function of $\theta$ satisfies:
\begin{equation}
	\begin{aligned}
		\log p(y_{1:T}; \theta) &= \log \left(p(y_1;\theta) \prod_{t=1}^{T-1} p(y_{t+1}|y_{1:t};\theta)\right)\\
		&= \log p(y_1;\theta) + \sum_{t=1}^{T-1} \log p(y_{t+1}|y_{1:t};\theta) \\
		&= \log \int p(y_1|x_1)p(x_1;\theta)\mathrm{d} x_1 + \sum_{t=1}^{T-1} \log \int p(y_{t+1}|x_{t+1}) p(x_{t+1}|y_{1:t};\theta) \mathrm{d}x_{t+1}.\\
	\end{aligned}
	\label{MDP-1}
\end{equation}
In the special case where $f$ and $h$ are linear, the marginal likelihood distributions $p(y_1;\theta)$ and $p(y_{t+1}|y_{1:t};\theta)$ have analytical solutions, and (\ref{MDP-1}) can be computed exactly. However, for nonlinear case, the exact computation of $p(y_1;\theta)$ and $p(y_{t+1}|y_{1:t};\theta)$ becomes intractable, necessitating some approximations to estimate (\ref{MDP-1}).

In the EnKF, the forecast distributions $p(x_1;\theta)$ and $p(x_{t+1}|y_{1:t};\theta)$ can be approximated by the forecast ensembles $\hat{x}_1^{1:N}$ and $\hat{x}_{t+1}^{1:N}$, respectively. Given $\theta$ and $y_{1:T}$, we can recursively obtain a trajectory consisting of the analysis and forecast ensembles according to the two steps (\ref{EnKF-fore}-\ref{EnKF-analysis}):
$$
	\tau_{\theta} = \{x_0^{1:N}, \hat{x}_1^{1:N}, x_1^{1:N},\hat{x}_2^{1:N}, \cdots, x_{T-1}^{1:N}, \hat{x}_T^{1:N}\}.
$$
Then the log-marginal-likelihood can be approximated by $\tau_{\theta}$ as follows:
\begin{equation}
	\begin{aligned}
		\log p(y_{1:T}; \theta)
		&\approx \sum_{t=0}^{T-1} \log \left(\frac{1}{N} \sum_{i=1}^{N} p(y_{t+1}|\hat{x}_{t+1}^{(i)})\right)\\
		&= \sum_{t=0}^{T-1} \log \left(\frac{1}{N} \sum_{i=1}^{N} \mathcal{N}(y_{t+1}; h(\hat{x}_{t+1}^{(i)}), R)\right),
	\end{aligned}
	\label{MDP-2}
\end{equation}
where $\mathcal{N}(\cdot|\mu,\Sigma)$ represents the probability density function of the Gaussian distribution with mean $\mu$ and covariance matrix $\Sigma$. This estimate depends on the trajectory $\tau_{\theta}$, and is inherently stochastic due to the randomness of $\tau_{\theta}$. The uncertainty in $\tau_{\theta}$ arises from three sources: (1) the randomness of  initial analysis ensemble $x_0^{1:N}$, (2) the process noise $\xi_t$ in the forecast step, and (3) the perturbed noise $\eta_{t+1}^{(i)}$ in the analysis step. To account for the randomness introduced by $\tau_{\theta}$, we take the expectation with respect to $\tau_{\theta}$ in (\ref{MDP-2}), resulting in the following estimate:
$$
	\log p(y_{1:T}; \theta)
	\approx \mathbb{E}_{\tau_{\theta}}\left[\sum_{t=0}^{T-1}\log \left(\frac{1}{N} \sum_{i=1}^{N} \mathcal{N}(y_{t+1}; h(\hat{x}_{t+1}^{(i)}), R)\right)\right].
$$
Therefore, computing the MLE of $\theta$ is approximated by the following problem:
\begin{equation}
	\theta^{*} = \arg \max_{\theta} \mathbb{E}_{\tau_{\theta}}\left[\sum_{t=0}^{T-1} \log \left(\frac{1}{N} \sum_{i=1}^{N} \mathcal{N}(y_{t+1}; h(\hat{x}_{t+1}^{(i)}), R)\right)\right].
	\label{objective}
\end{equation}

\subsection{Formulation of the MDP}
\label{MDP}
It can be observed that, if we define the reward function as
$$
r(s_t,a_t) = \log \left(\frac{1}{N} \sum_{i=1}^{N} \mathcal{N}(y_{t+1}; h(\hat{x}_{t+1}^{(i)}), R)\right),
$$
the objective in problem (\ref{objective}) takes a form similar to the RL objective in (\ref{RL-objective}) when $\gamma = 1$. This motivates us to reformulate the problem as an MDP and solve it within RL framework. A straightforward approach is to treat the analysis ensemble $x_t^{1:N}$ as a state and the forecast ensemble $\hat{x}_{t+1}^{1:N}$ as an action executed by the agent, with the state transition to $x_{t+1}^{1:N}$ determined by the analysis step of EnKF. However, note that $r(s_t,a_t)$ in this setting is dependent of the observation $y_{t+1}$, whereas in standard MDP, the reward function is only dependent of the state and action. Furthermore, the state transition process also depends on the observation $y_{t+1}$.  To address these discrepancies, we augment the state and action with the observation $y_{t+1}$, and subsequently construct the corresponding MDP for problem (\ref{objective}). Here, we provide a detailed description of the MDP designed for (\ref{objective}).

\textbf{State and action.} We incorporate the observation $y_{t+1}$ into the state and action spaces. Then the state and action can be represented as follows:
$$
\begin{aligned}
	s_t &= (x_{t}^{1:N}, y_{t+1}),\\
	a_t &= (\hat{x}_{t+1}^{1:N}, y_{t+1}).
\end{aligned}
$$

\textbf{Policy.} According to the definition of the state and action, the policy can be decomposed into two parts:
\begin{equation}
\begin{aligned}
	\pi_{\theta}(a_t|s_t) &= p_{\theta}(\hat{x}_{t+1}^{1:N}|x_{t}^{1:N})\cdot p(y_{t+1}|y_{t+1})\\
	&= \left(\prod_{i=1}^{N} p_{\theta}(\hat{x}_{t+1}^{(i)}| x_{t}^{(i)})\right) \cdot \delta_{y_{t+1}}(y),
\end{aligned}
\label{policy}
\end{equation}
where $\delta_{y_{t+1}}$ is the Dirac delta function at $y_{t+1}$, and the probability distribution $p_{\theta}(\hat{x}_{t+1}^{(i)}| x_{t}^{(i)})$ is determined by the surrogate state transition model (\ref{transition-model}) parameterized by $\theta$.

\textbf{State transition}. Note that $s_{t+1} = (x_{t+1}^{1:N}, y_{t+2})$.  Given $s_t$ and $a_t$, the ensemble $x_{t+1}^{1:N}$ can be obtained through the analysis step of the EnKF, and the component $y_{t+2}$ is directly available from the observation trajectory.

\textbf{Reward}. The reward function is defined as
$$
	r(s_t,a_t) = \log \left(\frac{1}{N} \sum_{i=1}^{N} \mathcal{N}(y_{t+1}; h(\hat{x}_{t+1}^{(i)}), R)\right).
$$
The objective of this MDP is to find the optimal policy parameters $\theta^{*}$ that maximize the expected discounted cumulative reward $J(\pi_{\theta})$:
\begin{equation}
	\begin{aligned}
		\theta^{*} = \arg \max_{\theta} J(\pi_{\theta}) = \arg \max_{\theta} \mathbb{E}_{s_0,a_0,\cdots,s_{T-1},a_{T-1}}\left[\sum_{t=0}^{T-1} \gamma^{t} r(s_t,a_t)\right].
	\end{aligned}
	\label{MLE-MDP}
\end{equation}
Since the observation $y_{t+1}$ in the state and action is directly specified by the given observation trajectory, the randomness of $\{(s_t,a_t)\}_{t=0}^{T-1}$ is
determined by the analysis and forecast ensembles, which is consistent with that of $\tau_{\theta}$. Therefore, the problem (\ref{MLE-MDP}) is equivalent to the problem (\ref{objective}) when $\gamma = 1$. From this perspective, the MLE problem can be reformulated as an MDP. Under this formulation, computing the MLE of $\theta$ is equivalent to finding the optimal policy parameters $\theta^{*}$ for the MDP, which can be efficiently solved using some RL techniques.

\section{RL based DA for unknown state model}
\label{RL-DA}
As discussed in the previous section, it is necessary to learn a surrogate state transition model (\ref{transition-model}) in the offline stage. In Section \ref{MDP-MLE}, we reformulated the process for computing the MLE of model parameters $\theta$ as an MDP. In this section, we introduce PPO, an efficient RL algorithm, as detailed in Subsection \ref{PPO-sec}. We then employ the PPO to learn the optimal policy parameters for the MDP, and present two corresponding algorithmic variants: PPO-EnKF and PPO-PF in Subsection \ref{PPO-EnKF} and Subsection \ref{PPO-PF}, respectively.

\subsection{PPO}
\label{PPO-sec}
PPO is a widely used RL algorithm designed to improve the stability and efficiency of policy gradient methods. It introduces a novel surrogate objective with clipped probability ratios, which significantly reduces computational complexity and simplifies implementation compared to the trust region policy optimization algorithm \cite{TRPO}. A detailed explanation of the PPO algorithm can be found in \cite{PPO}; here, we provide a brief overview.

PPO employs an actor–critic architecture, where both the actor and the critic can be parameterized by NNs. The actor represents the policy $\pi_{\theta}(a_t|s_t)$, which maps the current state $s_t$ to a probability distribution over possible actions. The critic $V_{\phi}(s_t)$, parameterized by $\phi$, estimates the state value function, which represents the expected discounted cumulative reward from state $s_t$ under the policy $\pi_{\theta}$.

In PPO, the maximization of $J(\pi_{\theta})$ is achieved by minimizing the actor loss $L_\text{actor}$ \cite{PPO1}, defined as
\begin{equation}
	L_\text{actor} = - \mathbb{E}_{t} \left[ \min \left(p_t(\theta) A_t, \text{clip}(p_t(\theta), 1-\epsilon,1+\epsilon)A_t \right)\right],
	\label{actor}
\end{equation}
where $p_t(\theta) = \frac{\pi_{\theta} (a_t|s_t)}{\pi_{\theta_{\text{old}}} (a_t|s_t)}$ represents the ratio between the new policy $\pi_{\theta}$ and the old policy $\pi_{\theta_{\text{old}}}$ used to generate the training data. The expectation $\mathbb{E}_t[\cdot]$ indicates the empirical average over a finite batch of samples. The advantage function $A_t$ quantifies how much better an action is compared to a baseline, often estimated using generalized advantage estimation (GAE) \cite{GAE}:
\begin{equation}
	\begin{aligned}
		A_t &= \delta_t + (\gamma \lambda)\delta_{t+1}+\cdots+(\gamma \lambda)^{T-t-1}\delta_{T-1},\\
		\delta_t &= r_t + V_{\phi}(s_{t+1}) - V_{\phi}(s_t),
	\end{aligned}
	\label{GAE}
\end{equation}
where $r_t := r(s_t,a_t)$, and the parameter $\lambda \in [0,1]$ controls the trade-off between bias and variance in advantage estimation. Here, we set $\lambda = 0.9$. The minimization of the actor loss (\ref{actor}) can be interpreted as updating the policy in the direction of actions that outperform the baseline (i.e., $A_t > 0$). To prevent large policy updates that could destabilize training, the clipping mechanism constrains $p_t(\theta)$ to the interval $[1-\epsilon,1+\epsilon]$. In this work, we set $\epsilon = 0.2$. 

The baseline used in (\ref{GAE}) is $V_{\phi}(s_t)$, which is learned by minimizing the critic loss:
\begin{equation}
	L_\text{critic} = \mathbb{E}_t \left[ \left( V_t^{\text{target}} - V_{\phi}(s_t) \right)^2 \right],
	\label{critic}
\end{equation}
where $V_t^{\text{target}}$ is the target value function computed using the agent's interactions with the environment (e.g., via bootstrapping or Monte Carlo estimates).

PPO is an on-policy algorithm. At each iteration, actions are sampled using the current policy to generate trajectories. The collected data is then used to perform multiple epochs of optimization on the actor and critic losses defined in (\ref{actor}) and (\ref{critic}). Through this iterative process, PPO gradually improves its ability to select optimal actions that maximize the objective function $J(\pi_{\theta})$.

\begin{figure}[htbp]
	\centering
	\includegraphics[scale=0.15]{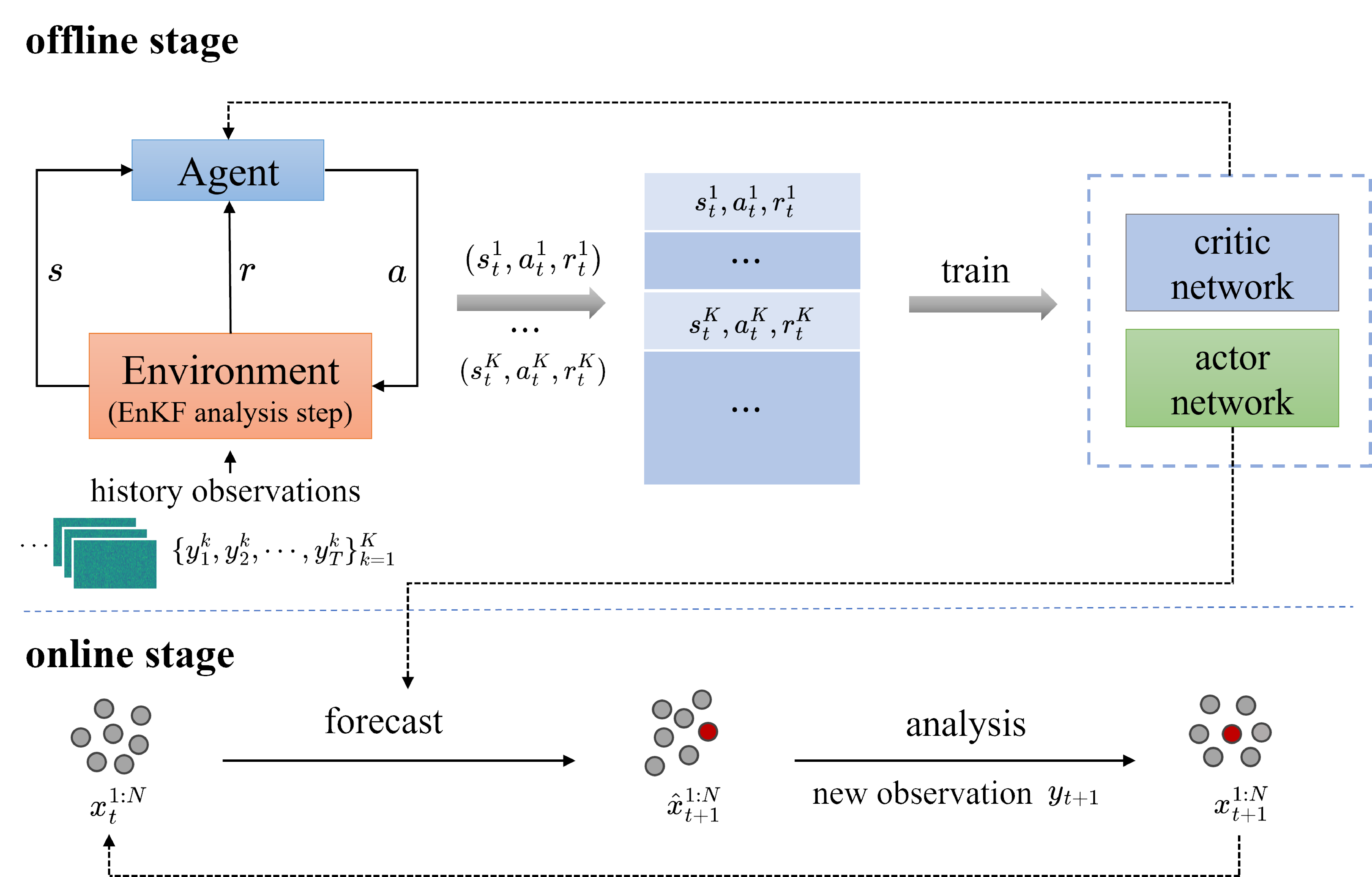}
	\caption{Workflow of PPO-EnKF.}
	\label{PPO-EnKF-fig}
\end{figure}

\subsection{PPO-EnKF}
\label{PPO-EnKF}
In the offline stage, we employ the PPO algorithm to learn the optimal model parameters $\theta^{*}$ for solving the optimization problem (\ref{MLE-MDP}). We begin by describing the NN architectures used for the actor and the critic. As shown in (\ref{policy}), the policy $\pi_{\theta}(a_t|s_t)$ can be decomposed into a probability distribution $p_{\theta}(\hat{x}_{t+1}^{(i)}| x_{t}^{(i)})$ and a known identical mapping. Therefore, it is sufficient to construct an NN to approximate $p_{\theta}(\hat{x}_{t+1}^{(i)}| x_{t}^{(i)})$, which is determined by the parameterized state transition model (\ref{transition-model}). This network consists of a mean network $F_\alpha^{\text{NN}}$ and trainable covariance matrix $Q_{\beta}$. In our numerical experiments, we employ different architectures for $F_\alpha^{\text{NN}}$ according to specific problem settings. Detailed configurations of the network architectures are provided in Appendix \ref{hyperparameter}.

For the critic, the state $s_t$ includes a set of particles $x_t^{1:N}$, which are treated as an order-invariant collection. To account for this symmetry, we adopt the following network architecture:
\begin{equation}
	V_{\phi}(s_t) = V_{\phi_3}\left(\sum_{i=1}^{N}V_{\phi_1}(x_t^{(i)}) + V_{\phi_2}(y_{t+1})\right),
	\label{critic-ppo-enkf}
\end{equation}
where $V_{\phi_1}, V_{\phi_2}, V_{\phi_3}$ are NNs. Specifically, $V_{\phi_1}$ extracts features from each individual particle, and the outputs are aggregated via average pooling to capture the global information of the ensemble. The network $V_{\phi_2}$ encodes the auxiliary input $y_{t+1}$, and $V_{\phi_3}$ integrates both the pooled particle features and the auxiliary representation to estimate the value function. This architecture effectively handles set-valued state representations and has been widely used in RL tasks involving ensemble-based inputs\cite{critic-net}.

In practice, multiple observation trajectories $\{y_{1:T}^{k}\}_{k=1}^{K}$ may be available from historical data, each initialized with different initial conditions, where $K$ denotes the number of trajectories. In this setting, $K$ parallel actors can be employed to collect interaction data for training. The offline training workflow of PPO-EnKF is illustrated in Algorithm \ref{PPO-EnKF-alg}. The algorithm begins by initializing the environment, along with the actor and critic networks. Subsequently, the $K$ parallel actors interact with the environment to generate data. Each agent receives a state, selects an action based on the current policy, and transitions to the next state according to the analysis step of the EnKF. The resulting state $s_t^k$, action $a_t^k$, and the corresponding reward $r_t^k$ are stored in the training dataset. The actor and critic networks are then updated over multiple epochs using the loss functions defined in (\ref{actor}) and (\ref{critic}). This process is repeated until the cumulative reward converges or shows no significant improvement. With the learned optimal policy parameters $\theta^{*}$, state estimation can be performed using the surrogate state transition model in the online stage. The complete process of PPO-EnKF is described in Figure \ref{PPO-EnKF-fig}.

\begin{rem}
	In the preceding content, we introduced the PPO-EnKF for learning the surrogate model (\ref{transition-model}) that maps $x_{t}$ to $x_{t+1}$. However, in many practical applications, the system may be influenced by external control inputs\cite{system-c}. In such cases, the state transition model can be represented as
	$$
	x_{t+1}= f (x_t,c_t) +\xi_t,\quad \xi_t \sim \mathcal{N}(0,Q),
	$$
	where $c_t$ represents the control input. Given the external inputs $c_{0:T-1}$ and noisy observations $y_{1:T}$, our goal is to learn a surrogate state transition model for the underlying dynamics. To account for the influence of control input, we augment the state $s_t$ by incorporating $c_t$, i.e., $s_t = (x_t^{1:N}, y_{t+1}, c_t)$ and thus learn the policy
	$$
	\begin{aligned}
		\pi_{\theta}(a_t|s_t) &= p_{\theta}(\hat{x}_{t+1}^{1:N}|x_{t}^{1:N}, c_t)\cdot p(y_{t+1}|y_{t+1})\\
		&= \left(\prod_{i=1}^{N} p_{\theta}(\hat{x}_{t+1}^{(i)}| x_{t}^{(i)}, c_t)\right) \cdot \delta_{y_{t+1}}(y)
	\end{aligned}
	$$
	using the PPO algorithm. This extended approach is referred to as PPO-EnKF with control (PPO-EnKFc). The critic network in PPO-EnKFc adopts a similar architecture with (\ref{critic-ppo-enkf}):
	$$
		V_{\phi}(s_t) = V_{\phi_3}\left(\sum_{i=1}^{N}V_{\phi_1}(x_t^{(i)}) + V_{\phi_2}(y_{t+1},c_t)\right).
	$$
\end{rem}

\begin{algorithm}[ht]
	\caption{the offline stage of PPO-EnKF}
	\hspace*{0.001in} {\bf Input:}
	observation trajectories $\{y_{1:T}^{k}\}_{k=1}^{K}$;\\
	\vspace{-15pt}
	\begin{algorithmic}[1]
		\State Initialization of the environment and networks;
		\For{iteration = $1,2,\ldots$}
		\State $\mathcal{D} = \{\}$;
		\For{k = 1,2,$\ldots$,K}		
		\State $s_0 = (x_{0}^{1:N,k}, y_{1}^k)$;
		\For {$t = 0,1,\ldots,T-1$}
		\State Sample from $\pi_{\theta}(a_t^k|s_t^k)$ to obtain $a_t^k$;
		\State Compute $s_{t+1}^k$ based on (\ref{EnKF-analysis}) of EnKF, and calculate $r_t^k$;
		\State $(s_t^k,a_t^k,r_t^k)$ are added to $\mathcal{D}$;
		\EndFor
		\EndFor
		\For{$j$ = $1:\text{epochs}$}
		\State Optimize the actor loss (\ref{actor}) w.r.t. $\theta$ on $\mathcal{D}$;
		\State Optimize the critic loss (\ref{critic}) w.r.t. $\phi$ on $\mathcal{D}$;
		\EndFor
		\EndFor
	\end{algorithmic}
	\hspace*{0.001in} {\bf Output:} the model parameters $\theta^{*}$
	\label{PPO-EnKF-alg}
\end{algorithm}

\subsection{PPO-PF}
\label{PPO-PF}
In the previous subsection, we approximate the observation log-marginal-likelihood function using the forecast particles obtained by EnKF. However, EnKF is an approximation method that lacks theoretical guarantees for accurately capturing the true forecast distribution in nonlinear setting. In this subsection, we adopt a similar framework but incorporate the PF, which enables provably accurate estimation of the true distribution. Based on this integration, we propose the corresponding PPO-PF method.

We firstly provide an overview of the PF \cite{PF-1}. In practice, PF is implemented using the following three-step procedure to propagate the filtering distribution from time $t$ to $t+1$:
\begin{itemize}
	\item \textbf{Forecast step.} Generate the forecast particle $\hat{x}_{t+1}^{(i)}$ through the state transition model:
	\begin{equation}
		\begin{aligned}
			&\hat{x}_{t+1}^{(i)} = f(x_{t}^{(i)}) + \xi_t^{(i)},\quad \xi_t^{(i)}\sim \mathcal{N}(0,Q),\\
			&p(x_{t+1}|y_{1:t}) \approx \frac{1}{N} \sum_{i=1}^{N} \delta_{\hat{x}_{t+1}^{(i)}}(x_{t+1}).\\
		\end{aligned}
		\label{PF-fore}
	\end{equation}
	
	\item \textbf{Analysis step.} Update the forecast distribution via Bayes’ formula:
	\begin{equation}
		p(x_{t+1}|y_{1:t+1}) \approx \sum_{i=1}^{N} w_{t+1}^{(i)} \delta_{\hat{x}_{t+1}^{(i)}}(x_{t+1}),
		\label{PF-analysis}
	\end{equation}
	where
	$$
	w_{t+1}^{(i)} = \tilde{w}_{t+1}^{(i)} / \left(\sum_{i=1}^N \tilde{w}_{t+1}^{(i)}\right),\quad \tilde{w}_{t+1}^{(i)} = p(y_{t+1}|\hat{x}_{t+1}^{(i)}).
	$$
	
	\item \textbf{Resampling.} A set of equally weighted particles $x_{t+1}^{1:N}$ is sampled from the weighted empirical distribution (\ref{PF-analysis}). This step is performed for mitigating particle degeneracy, where only a few particles have significant weights while the others have negligible contributions. After resampling, the filtering distribution is approximated by
	\begin{equation}
		p(x_{t+1}|y_{1:t+1}) \approx \sum_{i=1}^{N}  \delta_{x_{t+1}^{(i)}}(x_{t+1}).
		\label{PF-resample}
	\end{equation}
	
\end{itemize}

Similar to Section \ref{data-log}, we approximate $\log p(y_{1:T};\theta)$ using the forecast particles obtained by the PF, leading to the following estimate:
\begin{equation}
	\begin{aligned}
		\mathcal{L}_N^{\text{PF}}(\theta) := \mathbb{E}_{\tau_{\theta}^{\text{PF}}}\left[\sum_{t=0}^{T-1} \log \left(\frac{1}{N} \sum_{i=1}^{N} \tilde{w}_{t+1}\right)\right] = \mathbb{E}_{\tau_{\theta}^{\text{PF}}}\left[ \log \left(\prod_{t=0}^{T-1} \left(\frac{1}{N} \sum_{i=1}^{N} \tilde{w}_{t+1}\right)\right) \right],
	\end{aligned}
	\label{objective-PF}
\end{equation}
where $\tau_{\theta}^{\text{PF}}$ represents the trajectory consisting of the analysis and forecast ensembles recursively generated through the three steps (\ref{PF-fore}-\ref{PF-resample}) of the PF.

\begin{defn}
	(\cite{MCO}) Monte Carlo Objective (MCO). Let $\hat{p}_N(y)$ be an unbiased positive estimator of $p(y)$, i.e., $\mathbb{E}[\hat{p}_N(y)] = p(y)$. Then the Monte Carlo objective $\mathcal{L}_N(y,p)$ is defined as
	$$
	\mathcal{L}_N(y,p) = \mathbb{E}[\log \hat{p}_N(y)].
	$$
	Here, $N$ indexes the amount of computation needed to simulate $\hat{p}_N(y)$,  e.g., the number of particles.
\label{MCO}
\end{defn}
Let $\hat{p}_N(y_{1:T};\theta) := \prod_{t=0}^{T-1} \left(\frac{1}{N} \sum_{i=1}^{N} \tilde{w}_{t+1}\right)$. It is a well-known result that $\hat{p}_N(y_{1:T};\theta)$ is an unbiased estimator of $p(y_{1:T};\theta)$ \cite{MCO-1}. Therefore, according to Definition \ref{MCO}, $\mathcal{L}_N^{\text{PF}}(\theta)$ is an MCO, and satisfies the following properties.

\begin{prop}
(\cite{MCO}) Let $\mathcal{L}_N(y,p)$ be an MCO defined by an unbiased positive estimator $\hat{p}_N(y)$ of $p(y)$. Then,
\begin{itemize}
	\item [(1)] $\mathcal{L}_N(y,p) \leq \log p(y)$.
	\item [(2)] If $\log p(y)$ is is uniformly integrable and $\hat{p}_N(y)$  is strongly consistent, then $\mathcal{L}_N(y,p) \rightarrow \log p(y)$ as $N \rightarrow \infty$.
	\item [(3)] Let $g(N) = \mathbb{E}\left[(\hat{p}_N(y) - p(y))^6\right]$ be the 6th central moment. If the 1st inverse moment is bounded, i.e., $\lim \sup_{N\rightarrow \infty} \mathbb{E} \left[\hat{p}_N(y)^{-1}\right] < \infty$, then
	$$
	\log p(y) - \mathcal{L}_N(y,p) = \frac{1}{2} \text{var}\left(\frac{\hat{p}_N(y)}{p(y)}\right) + \mathcal{O}(\sqrt{g(N)}).
	$$
\end{itemize}
\end{prop}

From this proposition, we can deduce that $\mathcal{L}_N^{\text{PF}}(\theta)$ provides a lower bound for $\log p(y_{1:T};\theta)$. Furthermore, it has been shown in \cite{MCO-1} that $\hat{p}_N(y_{1:T};\theta)$ is a strongly consistent estimator. Hence, if $\log p(y_{1:T};\theta)$ is uniformly integrable, we have $\mathcal{L}_N^{\text{PF}}(\theta) \rightarrow \log p(y_{1:T};\theta)$ as $N\rightarrow \infty$.

Similar to PPO-EnKF, we can construct a specific MDP for maximizing $\mathcal{L}_N^{\text{PF}}(\theta)$. The key difference is that the state transition process of this MDP is implemented via the analysis and resampling steps (\ref{PF-analysis}-\ref{PF-resample}) of the PF. We also employ the PPO algorithm to learn the optimal policy for this MDP.

\section{Numerical results}
\label{results}
In this section, we present several numerical examples to demonstrate the effectiveness of the proposed methods. In Section \ref{cm}, we apply PPO-EnKF and PPO-PF to a 2-dimensional problem with highly nonlinear observation model, and compare our approaches with the EM-EnKF method \cite{EM-EnKS-2}. In Section \ref{Lorenz63}, we evaluate the performance of PPO-EnKF and PPO-PF on the Lorenz 63 system under varying levels of observation noise, highlighting its robustness in high-noise scenarios. Section \ref{Lorenz96} addresses a more challenging setting: a high-dimensional Lorenz 96 system with sparse measurements. In Section \ref{AC}, we consider an Allen-Cahn equation and its variant with external inputs, demonstrating the ability of our method to identify underlying dynamics in the presence of control inputs. The network architectures are detailed in Appendix \ref{hyperparameter}.

For each example, we generate a training set denoted by $D_{\text{train}} =\{y_{1:T_1}^{k}\}_{k=1}^{N_1}$, and a test set $D_{\text{test}} = \{x_{1:T_{2}}^{k}, y_{1:T_{2}}^{k}\}_{k=1}^{N_2}$. In the offline stage, the surrogate model is learned using $D_{\text{train}}$. During the online stage, we perform state estimation using filtering methods and compare posterior estimates with the ground truth. Furthermore, when real-time observations are unavailable, the surrogate model can be employed to perform multi-step predictions. To evaluate the performance of the proposed methods, we use the following metrics (the definitions are  provided in Appendix \ref{metrics}):
\begin{itemize}
	\item RMSE-a: measures the accuracy of point estimates by computing the root mean square error (RMSE) between the mean of the filtering distribution and the true state values
	
	\item{Continuous Ranked Probability Score (CRPS)}: evaluates the consistency between the cumulative distribution functions of the inferred state variables and the true state values
	
	\item RMSE-f(t): quantifies the forecast accuracy at a prediction horizon of $t$ time steps
\end{itemize}

Note that in each example, we use an explicit form of the state transition model to generate the training observation data. However, the proposed methods do not require prior knowledge of the transition model for their implementation.
\subsection{Uniform circular motion}
\label{cm}

\begin{figure}[ht!]
	\centering
	\subfigure[PPO-EnKF]{
		\includegraphics[scale = 0.435]{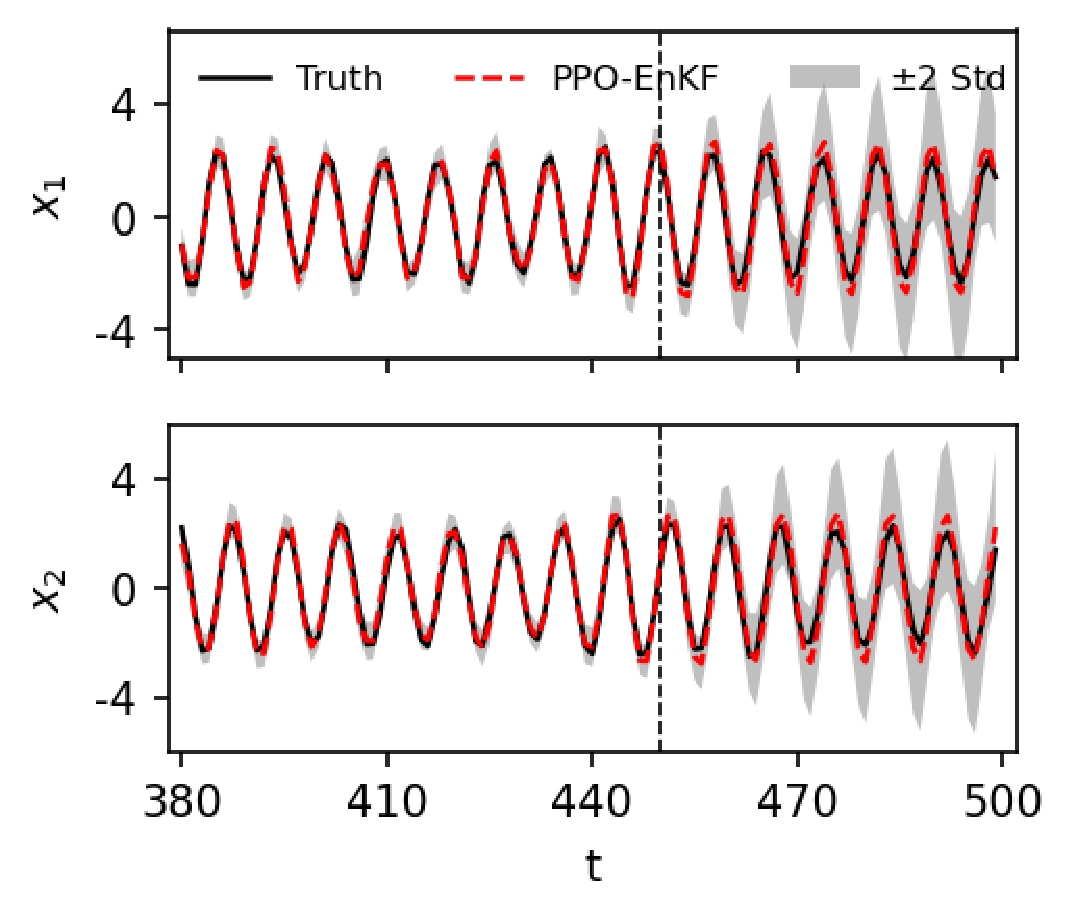}
	}
	\subfigure[PPO-PF]{
		\includegraphics[scale = 0.435]{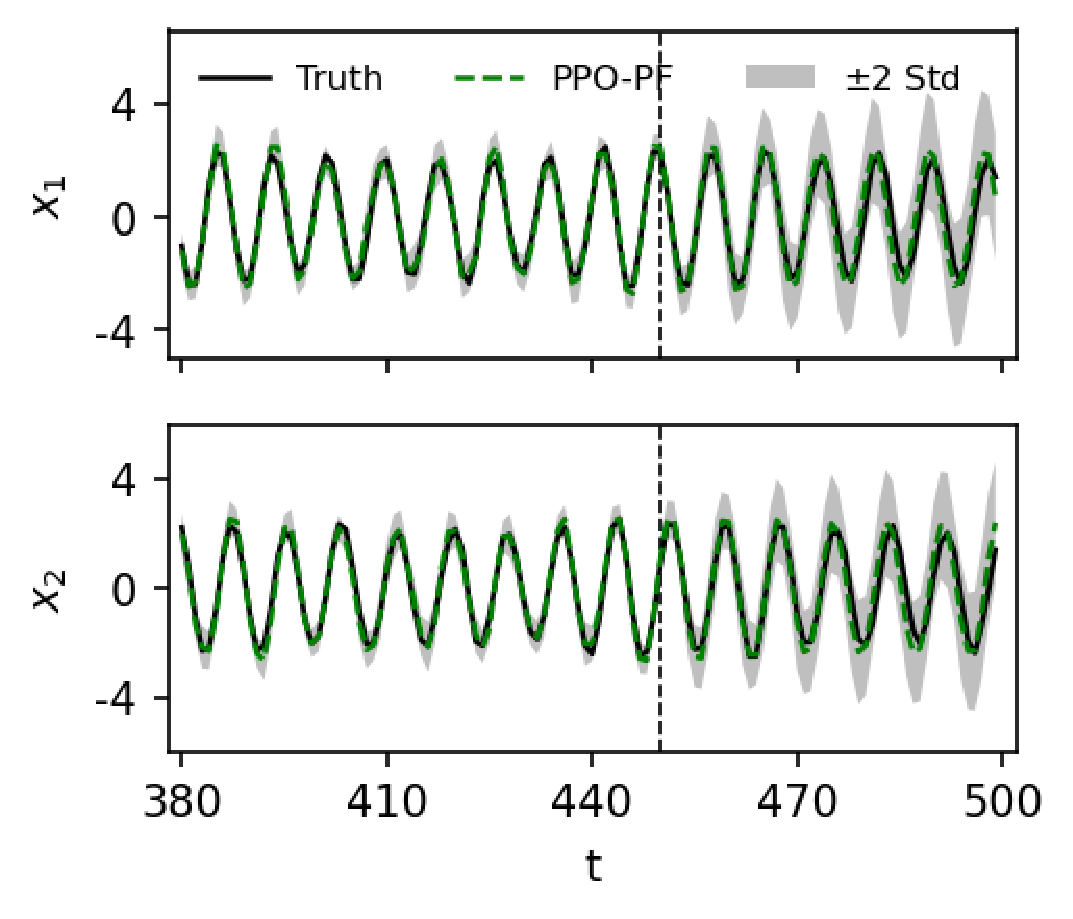}
	}
	\subfigure[EM-EnKF]{
		\includegraphics[scale = 0.435]{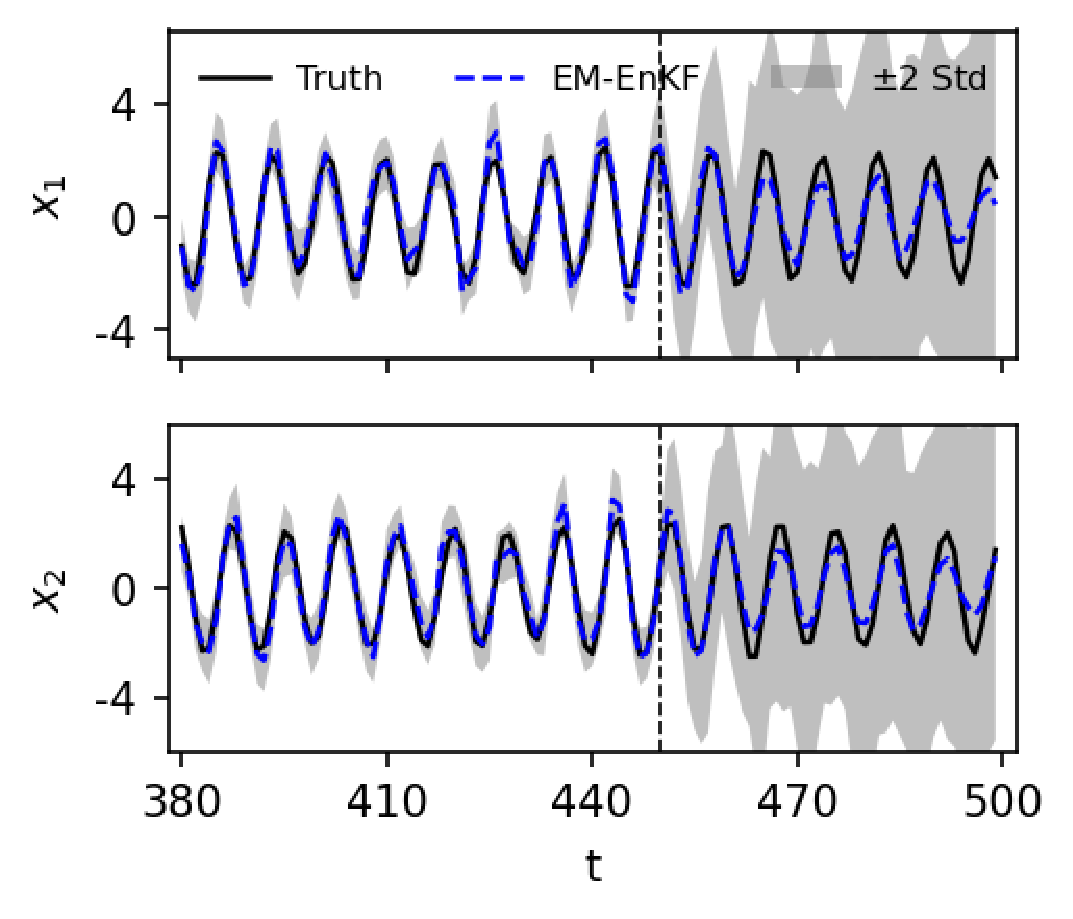}
	}
	\caption{State estimation and forecasting results using the linear observation function $h_1$ from time step $t = 380$ to $t = 500$, with estimation from $t = 380$ to $t = 450$ and forecasting from $t = 450$ to $t = 500$.}
	\label{ucm_tra_I}
\end{figure}

Consider a state evolution model that describes the uniform circular motion\cite{ucm} expressed by
\begin{equation}
	\begin{aligned}
		x_{t+1} &= \begin{bmatrix}
			\cos(\theta) & -\sin(\theta) \\
			\sin(\theta) &  \cos(\theta) \\
		\end{bmatrix} x_t + \xi_t,\quad \xi_t\sim \mathcal{N}(0, \sigma_x^2 I_2),\\
	\end{aligned}
	\label{suanli1}
\end{equation}
where $x_t = [x_{t,1}, x_{t,2}]^{\top}$ is the position vector and $\theta = \frac{\pi}{4}$ is the angular velocity. The process noise $\xi_t$ follows a Gaussian distribution with covariance $0.01 I_2$. In this experiment, we evaluate the performance of PPO-EnKF and PPO-PF under both linear and nonlinear observation models, and compare the results with EM-EnKF \cite{EM-EnKS-2}. In EM-EnKF, the ensemble Kalman smoother algorithm is employed to approximate the smoothing distribution in the offline stage, while EnKF is used for state estimation during the offline stage. The observation models are defined as follows:
\begin{equation}
	\begin{aligned}
		y_t &= h_1(x_t) + \eta_t = x_t + \eta_t,\quad \eta_t\sim \mathcal{N}(0, \sigma_y^2 I_2),\\
		y_t &= h_2(x_t) + \eta_t = \begin{bmatrix}
			\left\| x_t \right\|_2 \\
			\arcsin(\frac{x_{t,1}}{\left\| x_t \right\|_2})\\
			\arccos(\frac{x_{t,2}}{\left\| x_t \right\|_2})
		\end{bmatrix} + \eta_t,\quad \eta_t\sim \mathcal{N}(0, \sigma_y^2 I_2),
	\end{aligned}
	\label{suanli1}
\end{equation}
where $\left\| \cdot \right\|_2$ is the $l_2$-norm, and observation noise $\eta_t$ follows a Gaussian distribution with covariance $0.4 I_2$.

\begin{figure}[ht!]
	\centering
	\subfigure[PPO-EnKF]{
		\includegraphics[scale = 0.435]{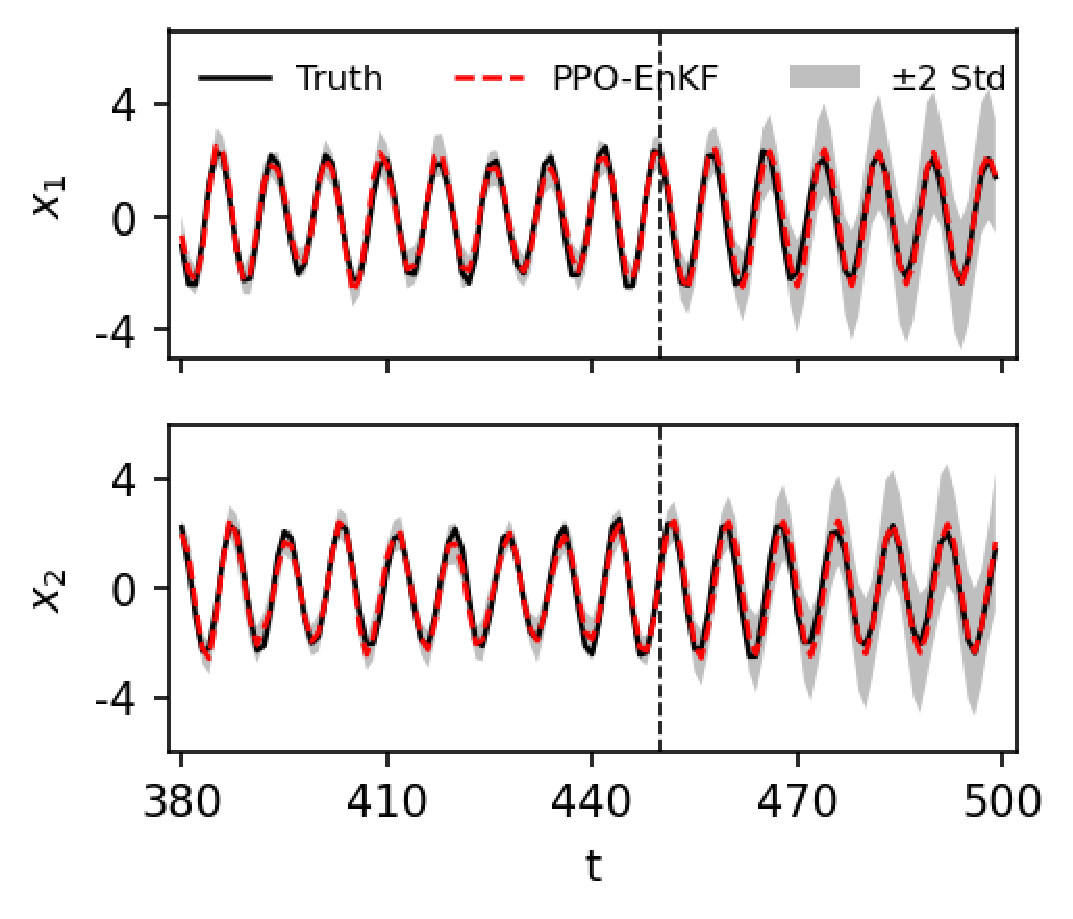}
	}
	\subfigure[PPO-PF]{
		\includegraphics[scale = 0.435]{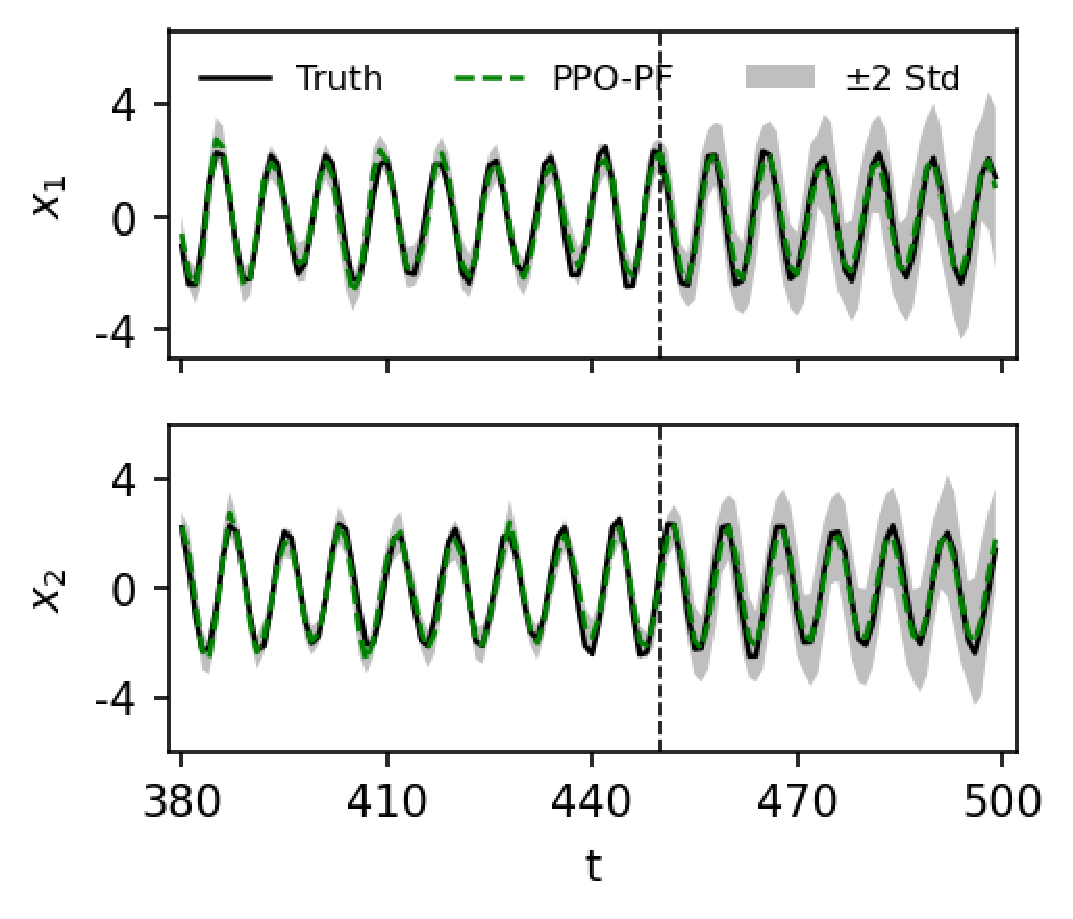}
	}
	\subfigure[EM-EnKF]{
		\includegraphics[scale = 0.435]{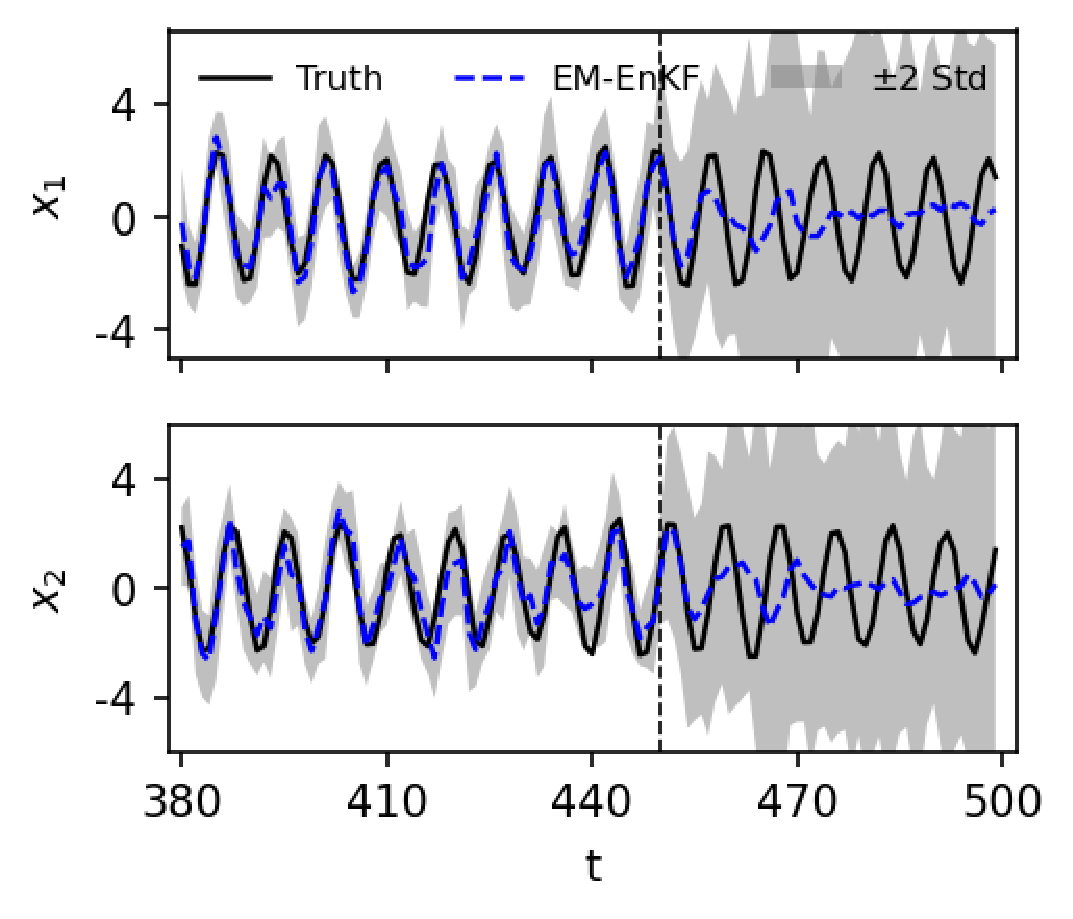}
	}
	\caption{State estimation and forecasting results using the nonlinear observation function $h_2$ from time step $t = 380$ to $t = 500$, with estimation from $t = 380$ to $t = 450$ and forecasting from $t = 450$ to $t = 500$.}
	\label{ucm_tra}
\end{figure}

We generate the training set with $T_1 = 400, N_1 = 20$ and the test set with $T_2 = 500, N_2 = 50$. The ensemble size is set to $N = 20$ for all methods. The state estimation is performed from time step $t = 0$ to $t = 450$ using the learned surrogate models, followed by forecasting from $t = 450$ to $t=500$ on the test set. Figure \ref{ucm_tra_I} and \ref{ucm_tra} present the state estimation and forecast results for a test trajectory from $t = 380$ to $t=500$ under two observation functions $h_1$ and $h_2$, respectively. We observe that the state can be accurately recovered by the PPO-EnKF and PPO-PF for both linear and nonlinear observations. In contrast, while EM-EnKF performs adequately under linear observations, its performance deteriorates in the nonlinear case, exhibiting some larger deviations in state estimation. Moreover, we find that the model learned by the EM-EnKF has higher model error, resulting in larger uncertainty in forecasting. Thus, its performance in long-term prediction is worse than the proposed methods. These results show that the RL framework enables the learning of more accurate surrogate models through trial-and-error interactions, demonstrating the effectiveness of the proposed methods for both state estimation and forecasting.

\subsection{Lorenz 63 model}
\label{Lorenz63}

\begin{figure}[ht!]
	\centering
	\subfigure[PPO-EnKF]{
		\includegraphics[scale = 0.365]{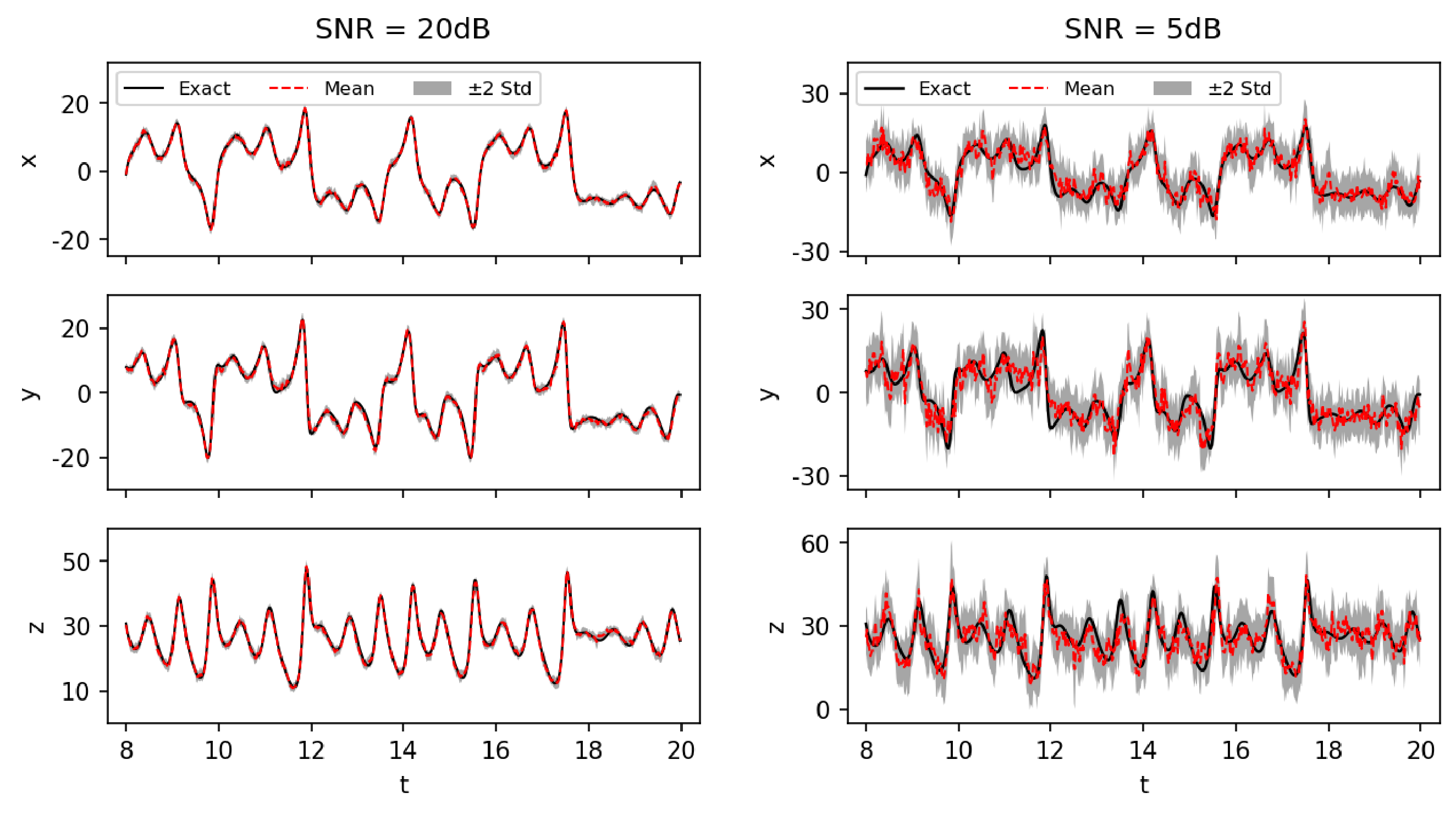}
	}
	\subfigure[PPO-PF]{
		\includegraphics[scale = 0.365]{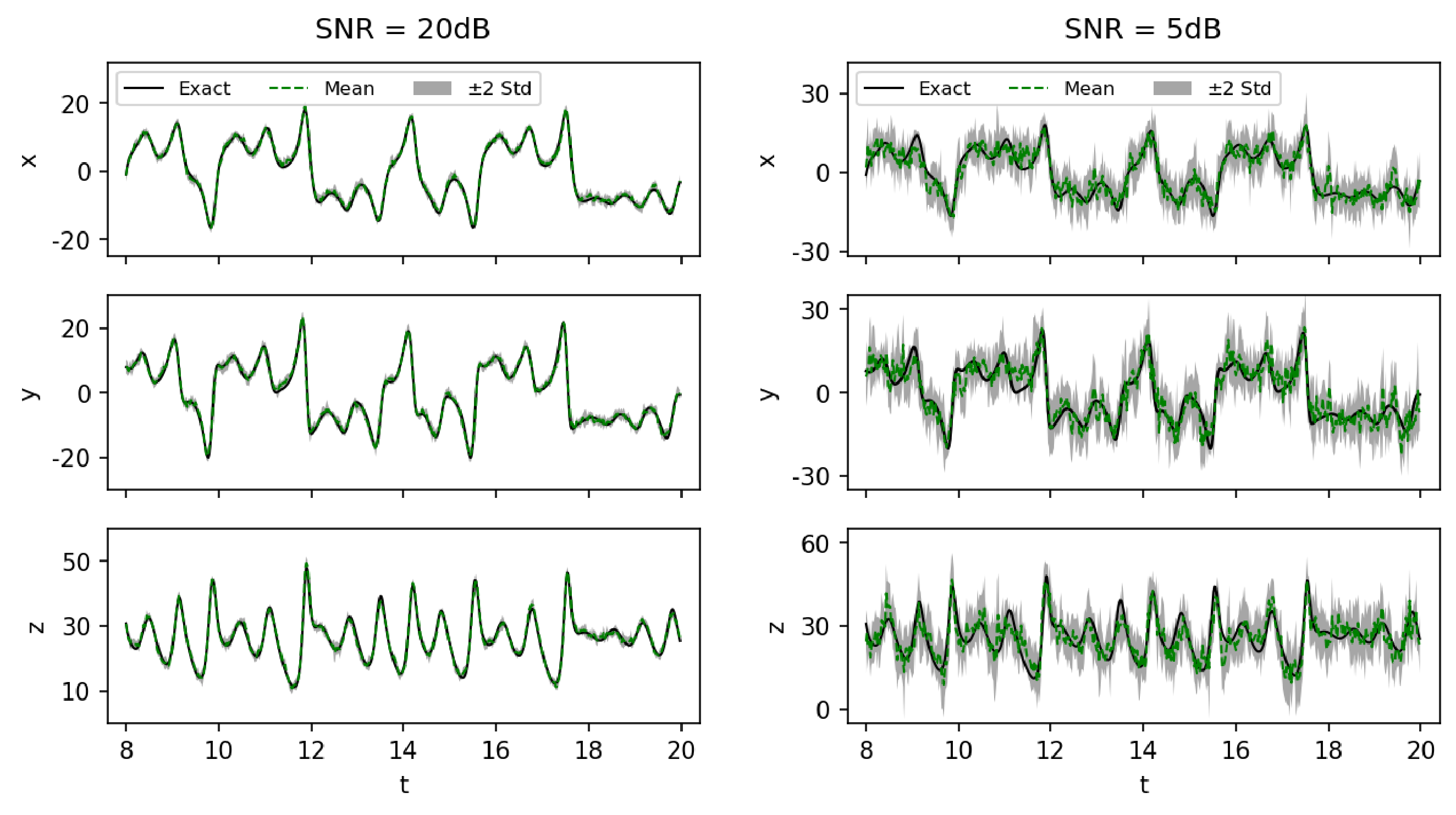}
	}
	\caption{The state estimation results obtained by PPO-EnKF and PPO-PF for a test trajectory under two different SNR conditions: $\mathrm{SNR} = 20 \mathrm{dB}$ and $\mathrm{SNR} = 5 \mathrm{dB}$ with the linear observation function $h_1$.}
	\label{63_tra_l}
\end{figure}

\begin{figure}[ht!]
	\centering
	\subfigure[PPO-EnKF]{
		\includegraphics[scale = 0.365]{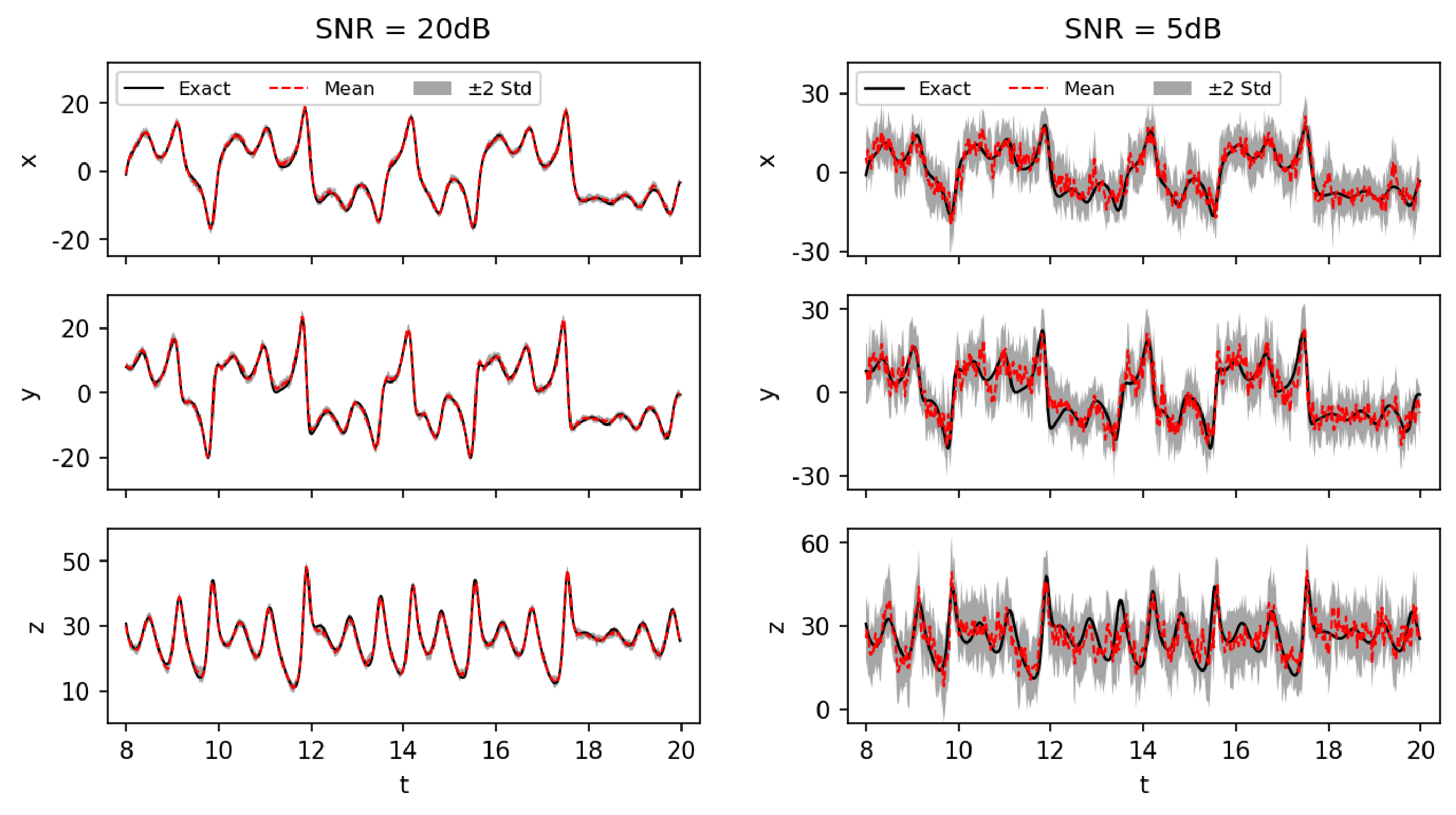}
	}
	\subfigure[PPO-PF]{
		\includegraphics[scale = 0.365]{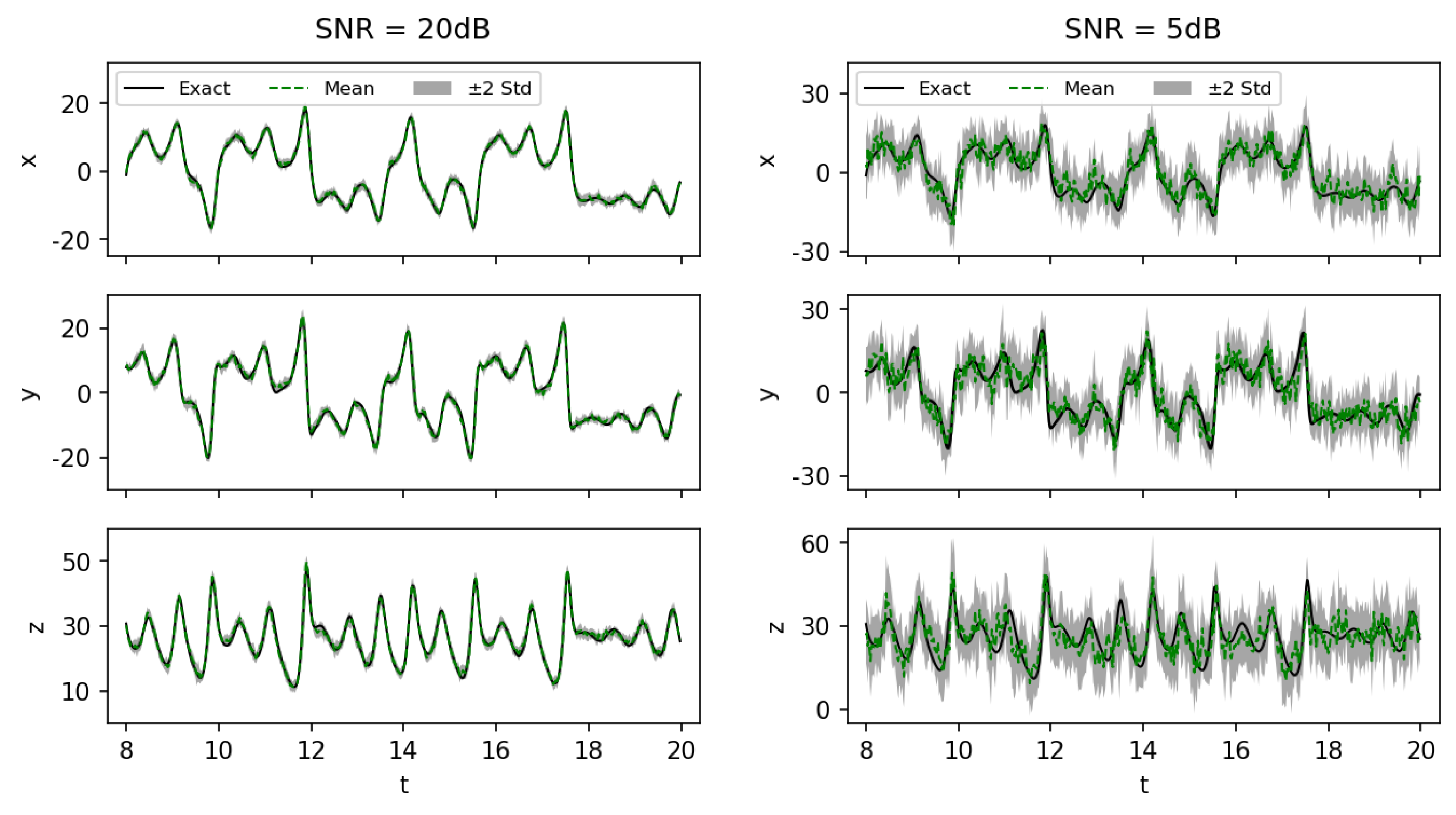}
	}
	\caption{The state estimation results obtained by PPO-EnKF and PPO-PF for a test trajectory under two different SNR conditions: $\mathrm{SNR} = 20 \mathrm{dB}$ and $\mathrm{SNR} = 5 \mathrm{dB}$ with the nonlinear observation function $h_2$.}
	\label{63_tra_nl}
\end{figure}

The Lorenz 63 is a simplified system developed to simulate atmospheric convection. It is governed by a set of three deterministic ordinary differential equations (ODEs):
\begin{equation}
	\begin{aligned}
		\frac{\mathrm{d} x_t}{\mathrm{d} t} &= \sigma (y_t-x_t), \\
		\frac{\mathrm{d} y_t}{\mathrm{d} t} &= x_t(\rho - z_t) - y_t, \\
		\frac{\mathrm{d} z_t}{\mathrm{d} t} &= x_t y_t - \beta z_t, \\
	\end{aligned}	
	\label{Lorenz_63}
\end{equation}
where $u_t = (x_t,y_t,z_t)$ represents the state vector. This system is known to exhibit chaotic behavior with the classical parameter setting $(\sigma, \rho, \beta) = \left(10,28, \frac{8}{3}\right)$. We simulate the ODEs (\ref{Lorenz_63}) using the Euler scheme with a time step $\Delta t = 0.02$. For the observation model, we consider both linear and nonlinear functions defined as
$$
\begin{aligned}
	y_t &= h_1(u_t) + \eta_t = u_t + \eta_t, \quad \eta_t\sim \mathcal{N}(0,\sigma_y^2 I_3),\\
	y_t &= h_2(u_t) + \eta_t = \begin{bmatrix}
		x_t + \frac{1}{2} \sin(x_t)\\
		y_t + \cos(z_t)\\
		y_t + z_t
	\end{bmatrix} + \eta_t, \quad \eta_t\sim \mathcal{N}(0,\sigma_y^2 I_3),
\end{aligned}
$$
where $\eta_t$ is the observation noise. The purpose of this experiment is to verify the effectiveness of the proposed methods under high levels of observation noise. We use the signal-to-noise ratio (SNR) to quantify the noise level.

\begin{figure}[ht!]
	\centering
	\includegraphics[scale=0.38]{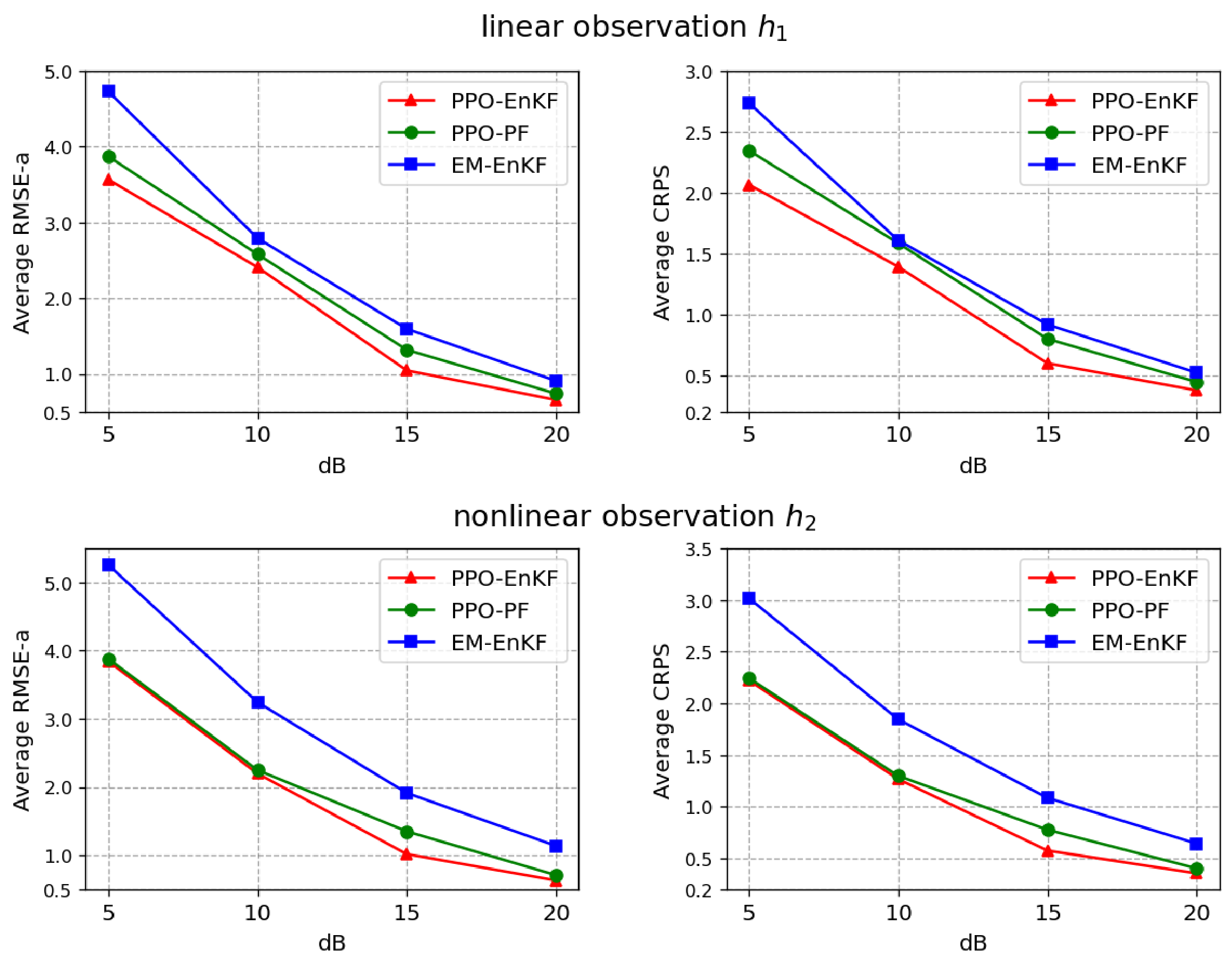}
	\caption{Performance metrics for state estimation across varying SNR levels.}
	\label{63-index}
\end{figure}

\begin{figure}[ht!]
	\centering
	\includegraphics[scale=0.4]{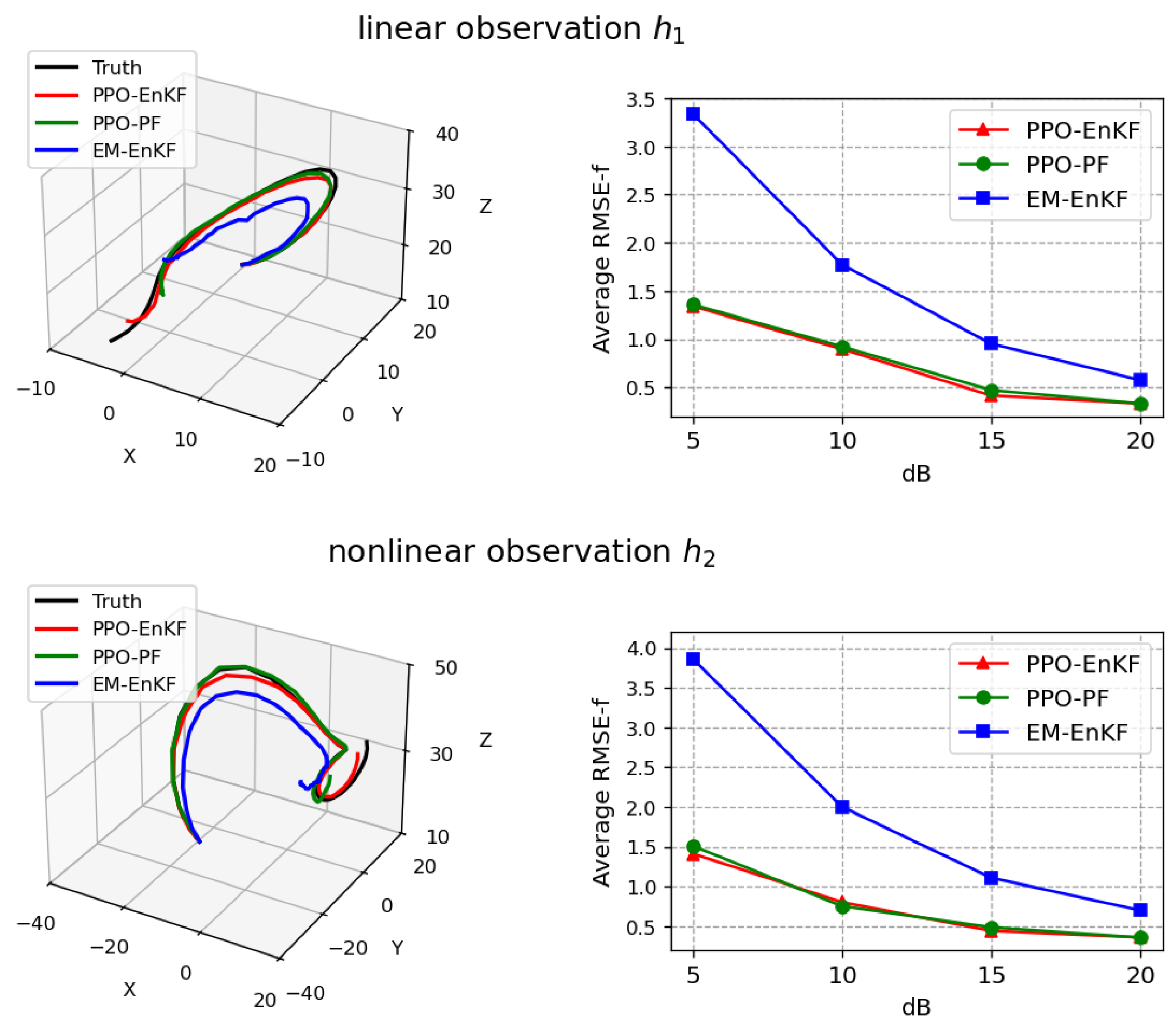}
	\caption{Left column: the sample mean predictions obtained from PPO-EnKF, PPO-PF and EM-EnKF with the same initial condition. Right column: RMSE-f(t) across varying SNR levels at a prediction horizon $t = 0.04$.}
	\label{63_fore}
\end{figure}

We generate the training set with $T_1=400, N_1=10$ and the test set with $T_{2}=1000, N_{2}=10$. The ensemble size is set to $N = 20$. Figure \ref{63_tra_l} and Figure \ref{63_tra_nl} present the state estimation results obtained by PPO-EnKF and PPO-PF for a test trajectory under two different SNR conditions: $\mathrm{SNR} = 20 \mathrm{dB}$ and $\mathrm{SNR} = 5 \mathrm{dB}$, with linear and nonlinear observation functions, respectively. We observe that when $\mathrm{SNR} = 20 \mathrm{dB}$, the filter estimates closely align with the true values. Even when the state is heavily corrupted by noise ($\mathrm{SNR} = 5 \mathrm{dB}$), PPO-EnKF and PPO-PF maintain robust estimation performance.

Furthermore, we evaluate two performance metrics: RMSE-a and CRPS across varying SNR levels and compare the results with EM-EnKF, as shown in Figure \ref{63-index}. The results indicate that PPO-EnKF and PPO-PF consistently outperform EM-EnKF across a range of noise conditions, with the performance gap being especially significant in the nonlinear case. This highlights the robustness and effectiveness of the proposed methods in handling nonlinear problems. Additionally, PPO-EnKF achieves higher filtering accuracy than PPO-PF, which may be attributed to the PF’s sensitivity to ensemble size. Since PF often requires a larger number of particles to perform well, its performance tends to degrade when the ensemble size is limited. In contrast, the EnKF exhibits more stable performance under such constraints.

To evaluates the forecasting performance of the surrogate models, we present the sample mean predictions obtained from the three methods with the same initial condition, as shown in Figure \ref{63_fore}. Additionally, the RMSE-f(t) at a prediction horizon $t = 0.04$ is also computed across varying SNR levels. Compared to EM-EnKF, the proposed methods demonstrate superior long-term tracking of the reference trajectory and achieve higher predictive accuracy across varying levels of observation noise.

\subsection{Lorenz 96 model}
\label{Lorenz96}
In this subsection, we evaluate the performance of the proposed methods on a high-dimensional dynamical system: Lorenz 96. This model is widely used to describe atmospheric circulation processes, which is governed by the following equations with periodic boundary conditions:
\begin{equation}
\begin{aligned}
	\frac{\mathrm{d} x_{t,j}}{\mathrm{d}t} &= x_{t,j-1}(x_{t,j+1} - x_{t,j-2}) - x_{t,j} +F, \quad j \in \{1,\cdots,m\},\\
	x_{t,0} &= x_{t,m}, \ x_{t,m+1} = x_{t,1}, \ x_{t,-1} = x_{t,m-1},
\end{aligned}
\label{Lorenz96-model}
\end{equation}
where $\mathbf{x}_t = \left(x_{t,1}, x_{t,2}, \cdots,x_{t,m}\right)$ represents an $m$-dimensional state vector and $F$ denotes an external forcing term. With a classical setting $F = 8$, the system exhibits chaotic behavior. In this experiment, we focus on the 40-dimensional Lorenz 96 system and discretize the governing equations using a fourth-order Runge-Kutta scheme with a time step $\Delta t = 0.05$.

\begin{figure}[ht!]
	\centering
	\includegraphics[scale=0.525]{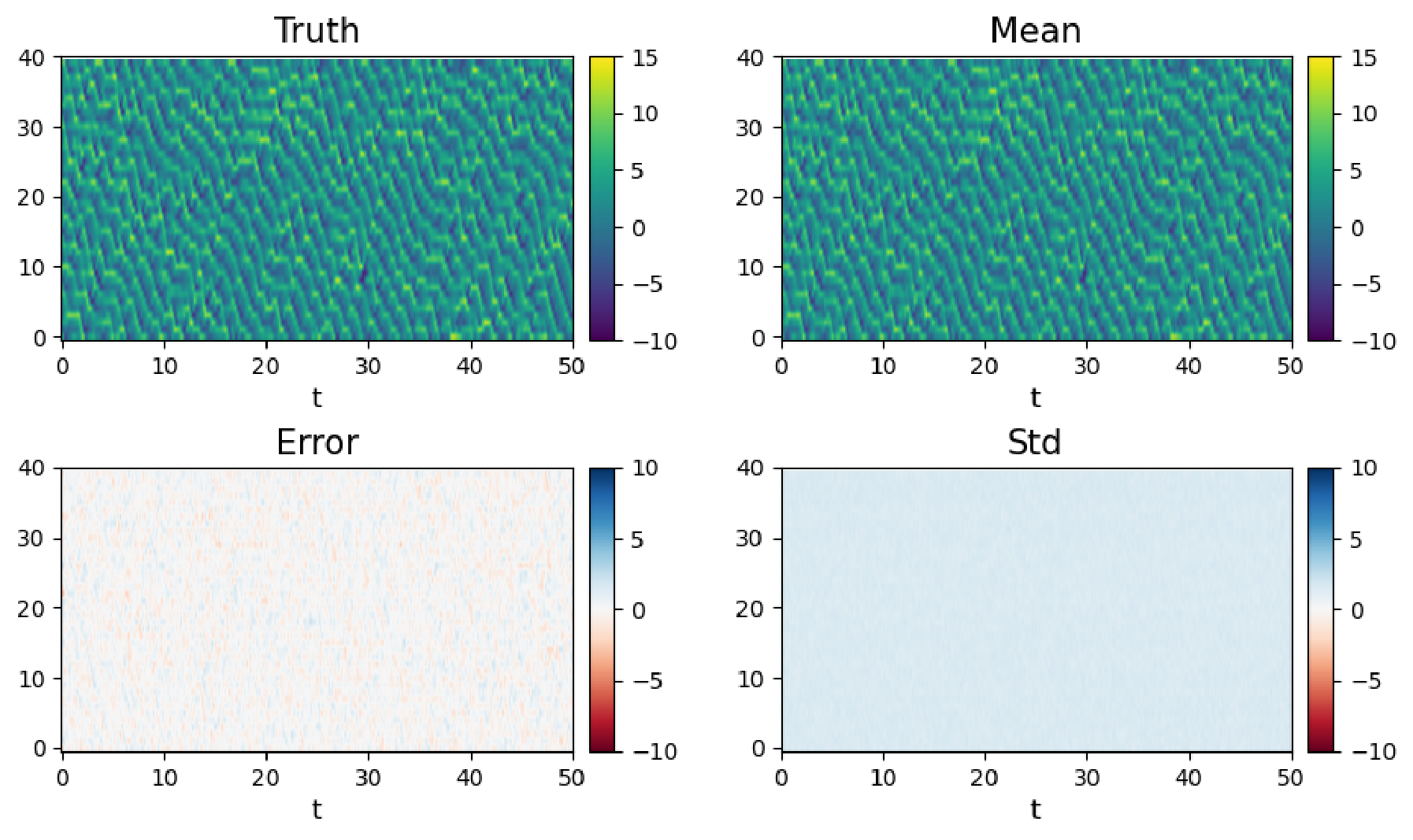}
	\caption{The state estimation results obtained by PPO-EnKF with identical observation function.}
	\label{96_state_40}
\end{figure}

\begin{figure}[ht!]
	\centering
	\includegraphics[scale=0.525]{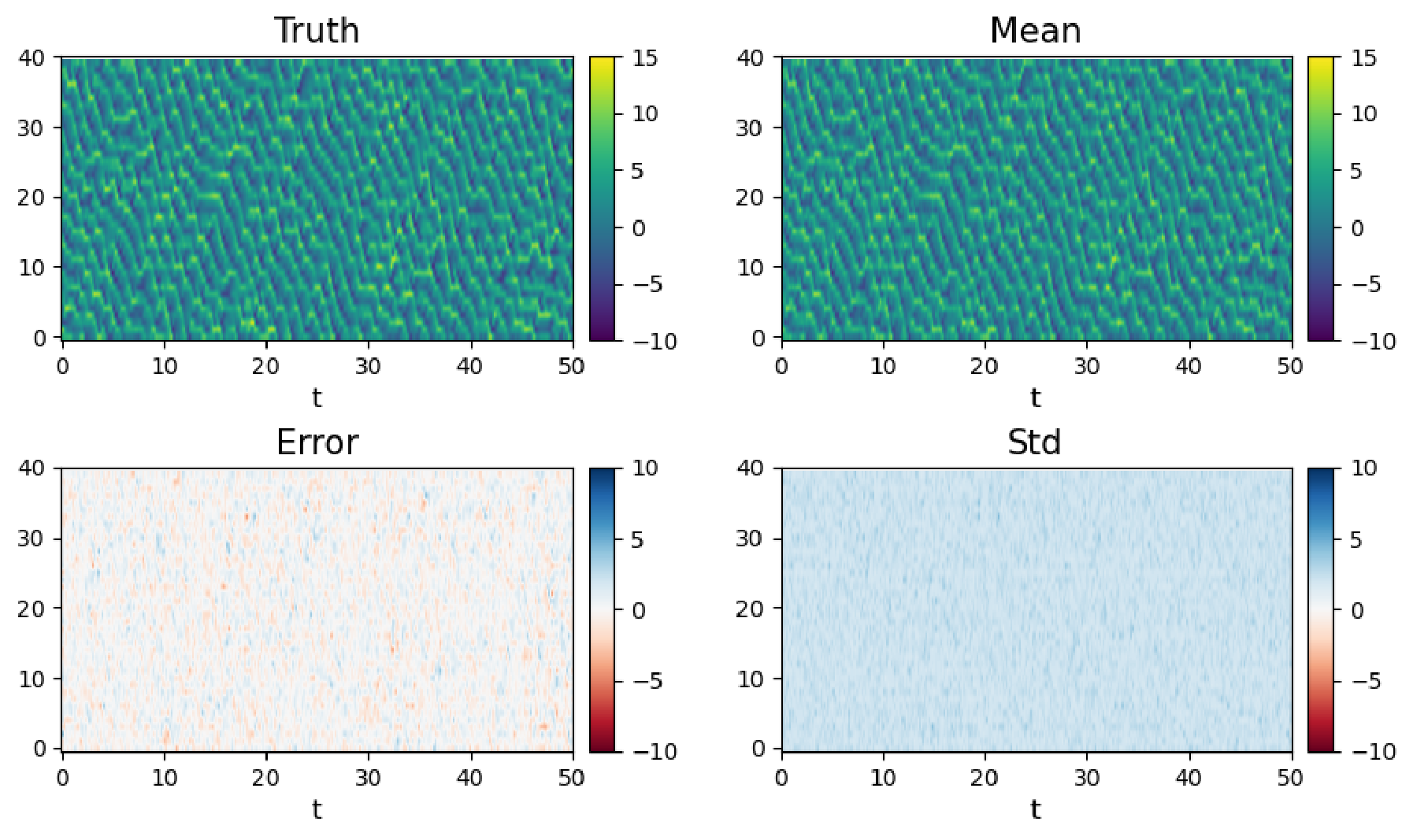}
	\caption{The state estimation results obtained by PPO-EnKF with sub-sampling function $\mathcal{H}_t^n$ ($n = 20$).}
	\label{96_state_20}
\end{figure}

We firstly consider an identical observation function with additive observation noise, represented as
$$
y_{t} = \mathbf{x}_t + \eta_t,
$$
where $\eta_t \sim \mathcal{N}(0,2 I_{40})$. For this high-dimensional chaotic system, we utilize a specific NN architecture in \cite{EM-EnKS-1} as the representation of $F_{\text{NN}}$, detailed in Appendix \ref{96-NN}. The training set consists of $N_1 = 4$ trajectories of length $T_1 = 1000$, and the test set comprises $N_{2}=5$ trajectories of $T_{2}=1000$. The ensemble size is set to $N = 50$, with all other settings consistent with the previous examples. The state estimation result obtained by PPO-EnKF is shown in Figure \ref{96_state_40}, demonstrating accurate performance with full observation information. In contrast, result for PPO-PF is not presented, as the filter exhibits divergence during the training phase in the absence of an accurate state transition model. This divergence is likely due to particle degeneracy, a known limitation of PF for high-dimensional problems. Therefore, PPO-EnKF is employed in the subsequent examples due to its superior robustness and reliability in handling high-dimensional problems.

\begin{figure}[ht!]
	\centering
	\includegraphics[scale=0.32]{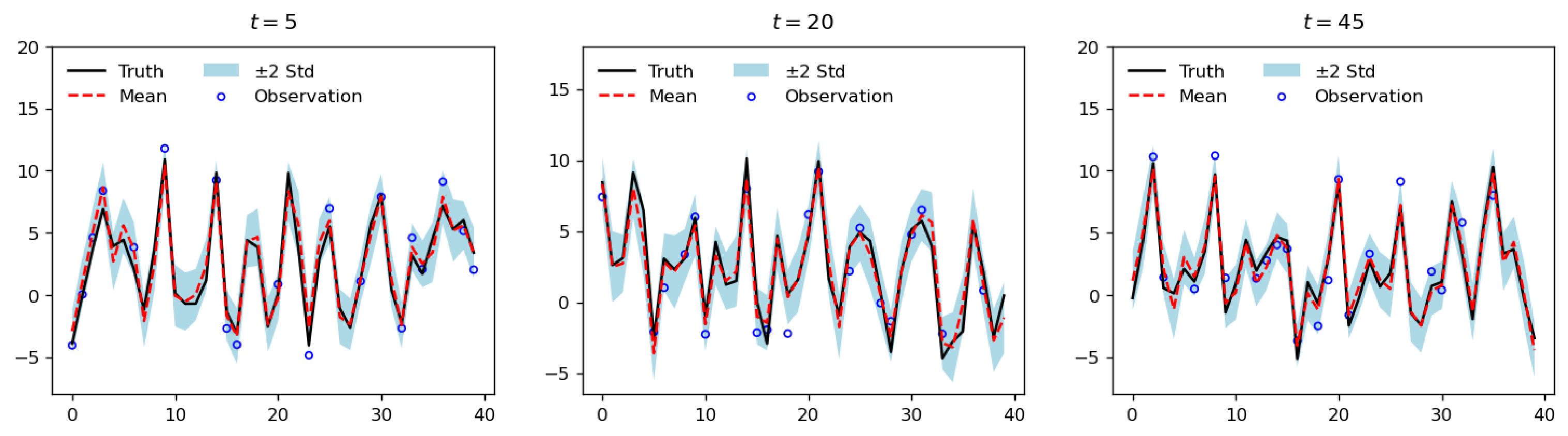}
	\caption{The estimated mean and uncertainty at some selected time points with sub-sampling function $\mathcal{H}_t^n$ ($n = 20$).}
	\label{96_state}
\end{figure}

\begin{figure}[ht!]
	\centering
	\includegraphics[scale=0.31]{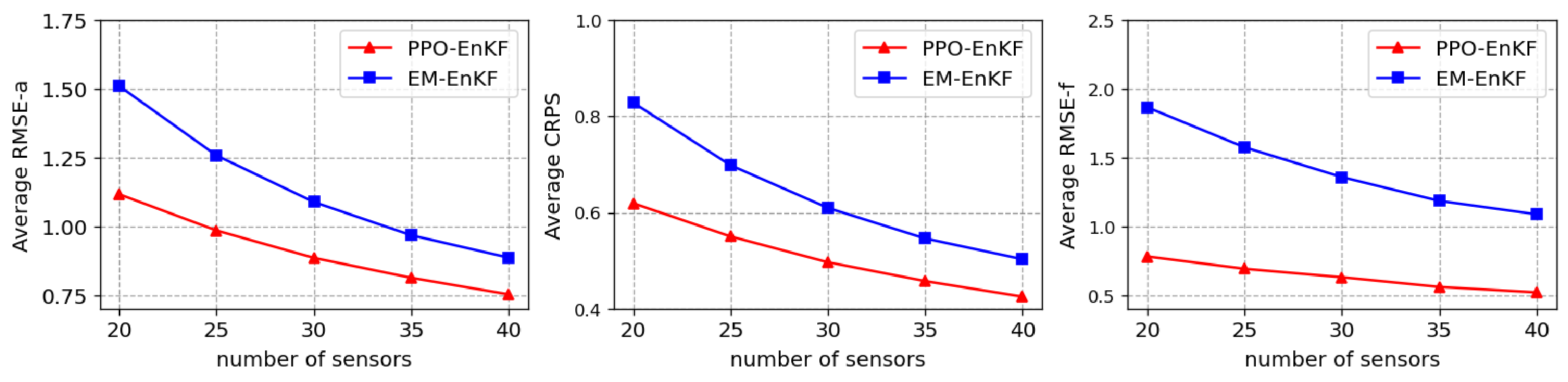}
	\caption{The three metrics: RMSE-a, CRPS, and RMSE-f($t = 0.2$) for varying numbers of sensors.}
	\label{96_metrics}
\end{figure}

The PPO-EnKF performs well in the full-observation setting. However, in many real-world scenarios, observations are often sparse and may be obtained from sensors with time-varying spatial locations. To assess the robustness of PPO-EnKF in dealing with such unstructured data, we consider a more challenging observation model, defined as follows:
$$
y_{t} = \mathcal{H}_t^n (\mathbf{x}_t) + \eta_t,
$$
where $\eta_t \sim \mathcal{N}(0,2 I_{40})$, and $\mathcal{H}_t^n$ represents a sub-sampling function that randomly selects $n$ sensor locations from a uniform distribution. This means that at each time step, only $n$ components of $\mathbf{x}_t$ are observed, with the sensor locations changing randomly over time. The result for $\mathcal{H}_t^n$ ($n = 20$) is presented in Figure \ref{96_state_20}. To further visualize the estimation accuracy across all state components, Figure \ref{96_state} displays the estimated mean and uncertainty at some selected time points. It can be observed that the posterior mean closely aligns with the reference even for the unobserved components, and the corresponding uncertainty estimates generally encompass the absolute errors.

To examine the effects under different sensor numbers, we evaluate the performance metrics: RMSE-a, CRPS, and RMSE-f ($t = 0.2$) for varying numbers of sensors and compare the results with EM-EnKF in Figure \ref{96_metrics}. Although PPO-EnKF's performance degrades as the number of observed components decreases, it consistently outperforms EM-EnKF, demonstrating its robustness in scenarios with limited observational data.

\subsection{Allen-Cahn equation}
\label{AC}

In this experiment, we evaluate the performance of PPO-EnKF and PPO-EnKFc on the Allen–Cahn equation under both uncontrolled and controlled settings. We firstly consider the uncontrolled Allen–Cahn equation with periodic boundary conditions, given by
$$
\begin{cases}
	\frac{\partial u}{\partial t} = \epsilon^2 \frac{\partial^2 u}{\partial x^2} + \mu (u - u^3), \quad x\in [-1,1], t\in [0,2], \\
	u(-1,t) = u(1,t),\\
	u_x(-1,t) = u_x(1,t),\\
	u(x,0) = u_0(x),
\end{cases}
$$
where $u_0(x)$ represents the initial condition. The parameters are set to $\epsilon = 0.001$ and $\mu = 3$. The Allen–Cahn equation \cite{A-C} is a classical model in the study of reaction–diffusion systems, with widespread applications in crystal growth \cite{A-C-growth}, phase transitions \cite{A-C-flow}, and image analysis \cite{A-C-image}. To numerically solve the system, we employ an implicit finite difference scheme on a uniform spatial-temporal grid with 40 spatial points and 200 time steps. The resulting discrete solutions $u_t$ are used to generate observations via a nonlinear observation model:
\begin{equation}
	\begin{aligned}
		y_t &= \arctan (u_t) + \eta_t,\quad \eta_t \sim \mathcal{N}(0,\sigma_y^2 I_{40}),
	\end{aligned}
	\label{A-C-observation}
\end{equation}
where the noise level is set to $\sigma_y = 0.1$. The initial condition $u_0(x)$ is randomly generated according to
$$
u_0(x) = U x^2 \cos(\pi x),\quad U \sim \text{Uniform} (0.8,1.2).
$$
We generate the training set $N_1 = 10, T_1 = 200$, and the test set $N_2 = 10, T_2 = 200$. The ensemble size is set to $N = 20$ for training.

\begin{figure}[ht!]
	\centering
	\includegraphics[scale=0.525]{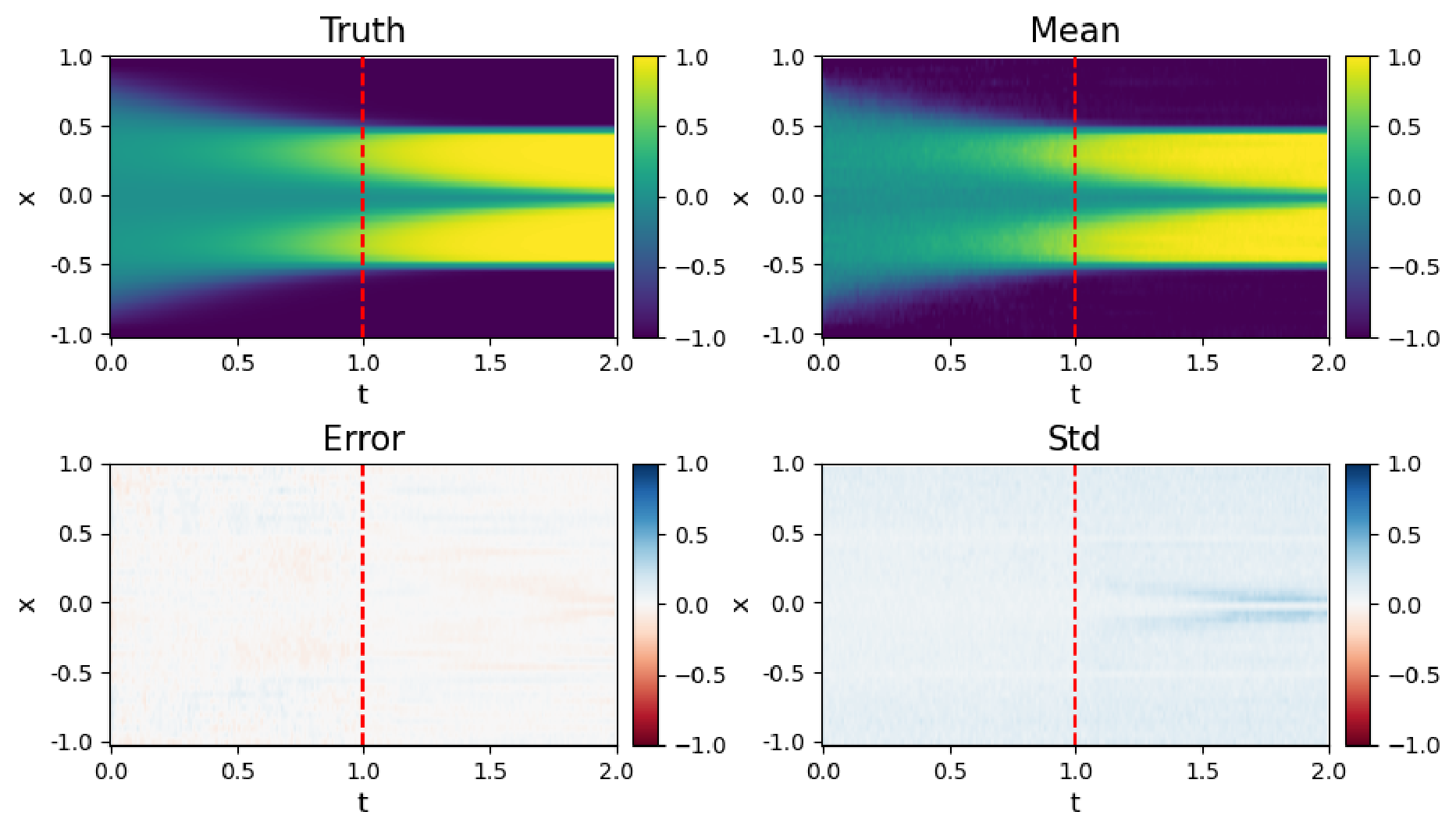}
	\caption{The state estimation and forecasting results for a test trajectory using PPO-EnKF. State estimation is performed over $[0,1]$, while forecasting is realized over $[1,2]$.}
	\label{AC-fig}
\end{figure}

\begin{figure}[ht!]
	\centering
	\includegraphics[scale=0.33]{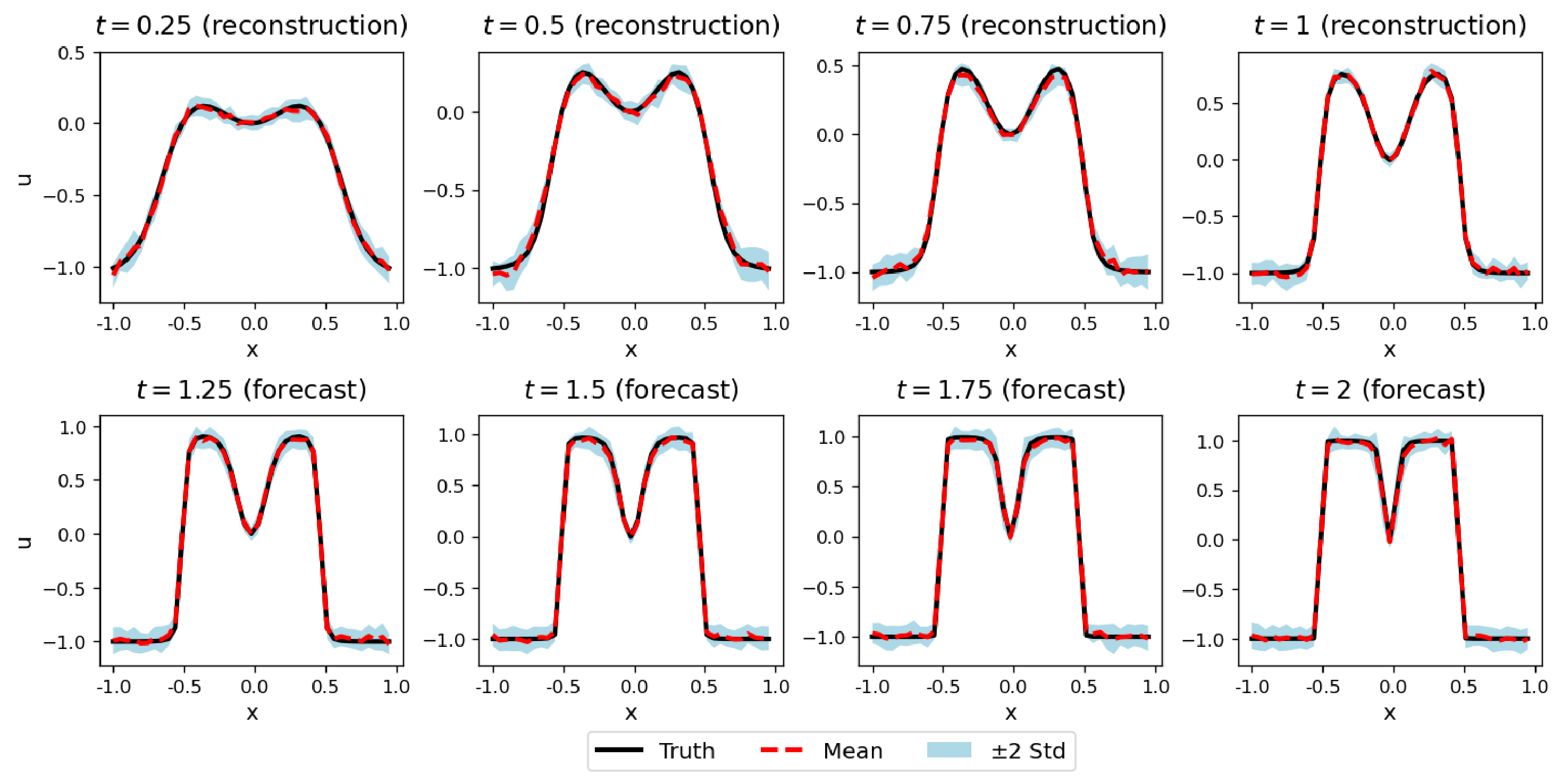}
	\caption{The estimated or predicted sample mean and uncertainty at some selected time points for the uncontrolled Allen-Cahn equation.}
	\label{AC-u-fig}
\end{figure}

\begin{figure}[ht!]
	\centering
	\includegraphics[scale=0.525]{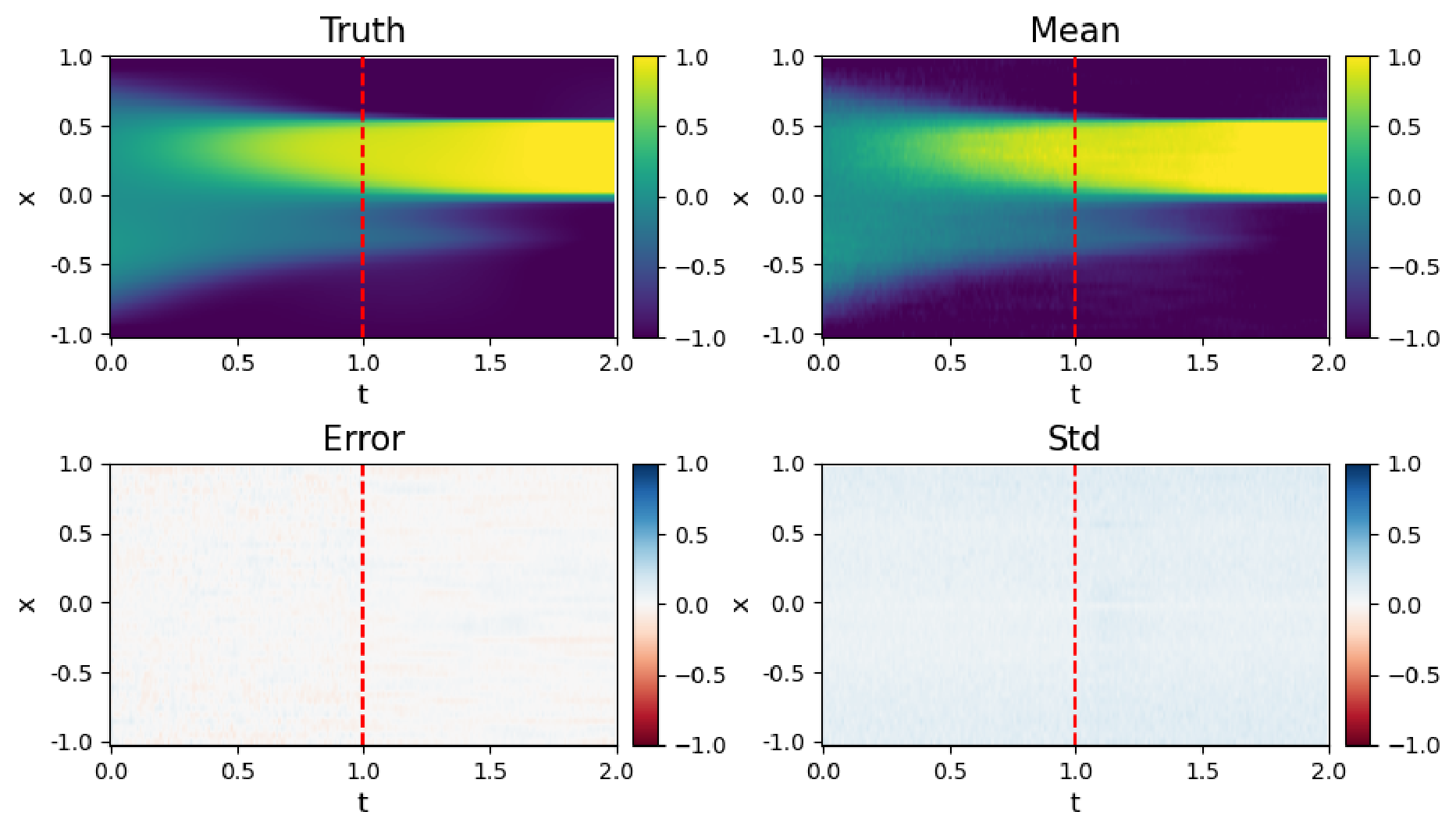}
	\caption{The state estimation and forecasting results for a test trajectory using PPO-EnKFc. State estimation is performed over $[0,1]$, while forecasting is realized over $[1,2]$.}
	\label{AC-c-fig}
\end{figure}

\begin{figure}[ht!]
	\centering
	\includegraphics[scale=0.33]{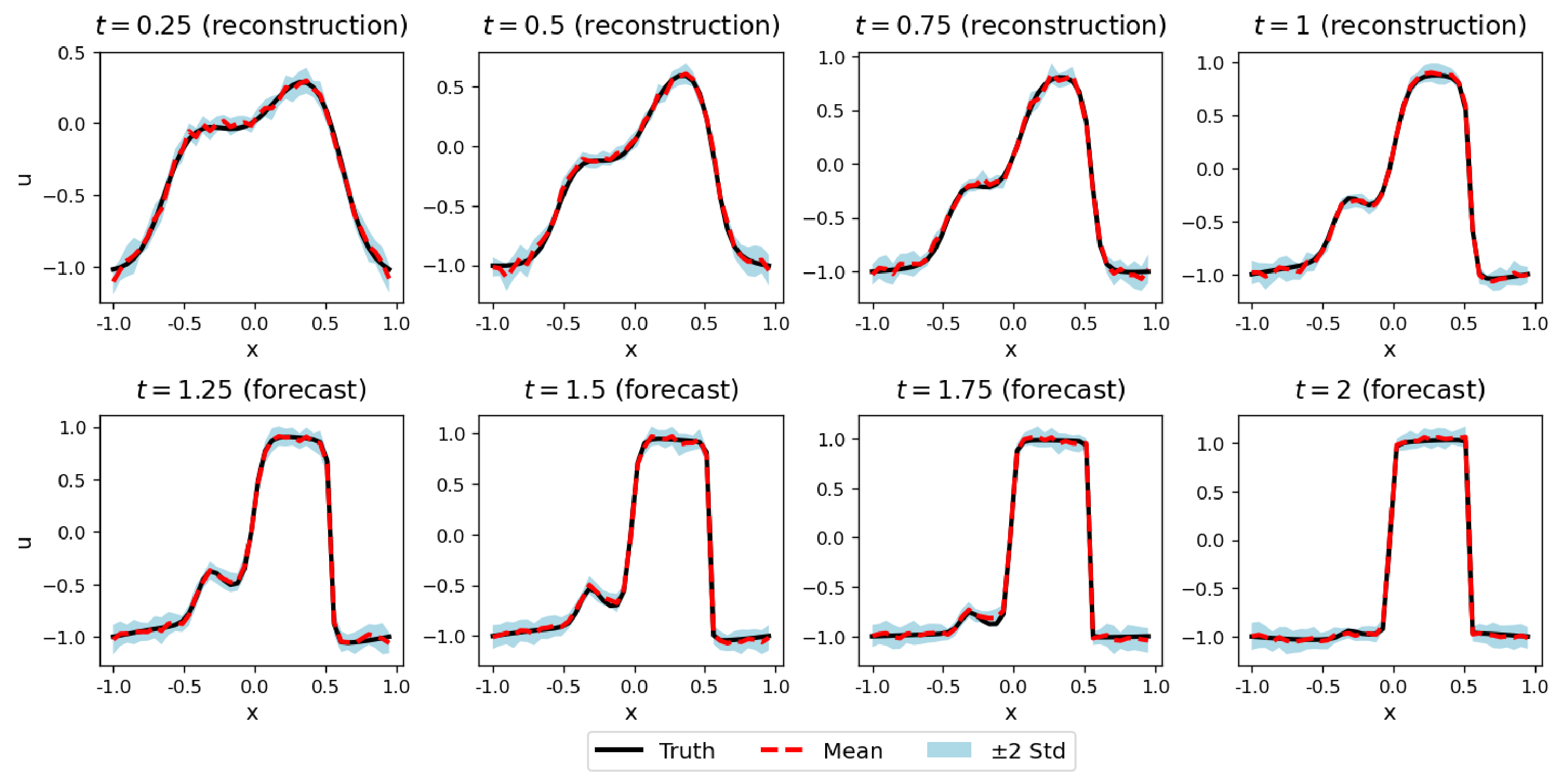}
	\caption{The estimated or predicted sample mean and uncertainty at some selected time points for the Allen-Cahn equation with control inputs.}
	\label{AC-u-c-fig}
\end{figure}

Figure \ref{AC-fig} presents the state estimation and forecasting results for a test trajectory using PPO-EnKF. State estimation is performed over the interval $[0,1]$, while forecasting is realized over $[1,2]$. Additionally, Figure~\ref{AC-u-fig} illustrates the estimated or predicted sample mean and uncertainty at some selected time points. The results demonstrate that PPO-EnKF is capable of providing reliable state estimation and accurate mean prediction for the uncontrolled system.

Then we consider the Allen–Cahn equation with a control \cite{A-C-control} input:
$$
\frac{\partial u}{\partial t} - \epsilon^2 \frac{\partial^2 u}{\partial x^2} - \mu (u - u^3) = a (x, t),
$$
where $a(x,t)$ represents a distributed control input and is  defined by
\[
a(x,t) = U_c \sin(\pi x) \cos(\pi t),\quad U_c \sim \text{Uniform}(0.4,0.6).
\]
To assess the effectiveness of PPO-EnKFc in capturing controlled dynamics, we generate a training dataset consisting of $N_1 = 10$ trajectories, each with $T_1 = 200$ time steps. The test dataset comprises $N_2 = 10$ trajectories with $T_2 = 200$ time steps. The observation model follows the same form as (\ref{A-C-observation}). The results for state estimation and forecasting are illustrated in Figure \ref{AC-c-fig} and Figure \ref{AC-u-c-fig}. It can be observed that PPO-EnKFc successfully captures the evolution of the controlled state variables from nonlinear observations.

\section{Conclusions}
\label{conclusions}
In this paper, we introduced the PPO-EnKF and PPO-PF for DA in scenarios where the state transition model is unknown. We showed that learning a surrogate state transition model from observations is equivalent to finding an optimal policy for a specific MDP, which can be effectively solved within RL framework through trial-and-error exploration. Numerical experiments have indicated that our proposed methods perform well for both linear and nonlinear observations, and demonstrated   good  accuracy and robustness.

 There are some questions required  for further investigation in future work. Firstly, in this study, we construct the surrogate model and apply the filtering methods in the original state space. Some model reduction techniques can be employed to improve the state estimation  for high-dimensional problems. Secondly, alternative ensemble-based filtering methods can be integrated into the proposed framework. Finally, our current approach applies RL to DA for learning unknown dynamics directly from noisy observations. Moreover, recent study explore the use of DA techniques to enhance the effectiveness of RL methods in partially observable MDPs \cite{DA-RL}. Future research could further investigate the integration of RL and DA to address complex and realistic problems.

\smallskip
\bigskip
\textbf{Acknowledgement:}
L. Jiang acknowledges the support of NSFC 12271408.

\bibliographystyle{elsarticle-num}
\bibliography{refs}

\begin{appendices}
	\section{Details for computing the metrics}
	\label{metrics}
	\subsection{RMSE-a}
	The RMSE-a quantifies the accuracy of point estimates by measuring the deviation between the posterior mean obtained from the filtering methods and the true state values $x_t^{\text{truth}}$:
	$$
	\text{RMSE-a} = \sqrt{\frac{1}{m T}\sum_{t=1}^{T} \left\| \frac{1}{N} \sum_{i=1}^{N} x_t^{(i)} - x_t^{\text{truth}} \right\| _2^2},
	$$
	where $m$ is the state dimensionality, and $T$ represents the total number of time steps.
	
	\subsection{CRPS}
	The CRPS is a metric in state estimation for assessing the difference between the cumulative distribution of posterior samples and the empirical distribution of the true state. For the $j$-th component of
	the state at time step $t$, it is defined as
	$$
	\text{CRPS}_{t, j} = \int_{-\infty}^{+\infty} \left(\mathbf{1}(x_{t,j}^{\text{truth}} < x) - \frac{1}{N} \sum_{i=1}^{N} \mathbf{1}(x_{t, j}^{(i)} < x)\right)^2 \mathrm{d}x,
	$$
	where $x_{t, j}^{(i)}$ represents the $j$-th component of the $i$-th sample $x_{t}^{(i)}$, $x_{t, j}^{\text{truth}}$ represents the $j$-th component of the true state  $x_{t}^{\text{truth}}$, and $\mathbf{1}(\cdot)$ is the indicator function. To evaluate the overall filtering performance, the CRPS is computed by averaging $\text{CRPS}_{t, j}$ across all state components and time steps:
	$$
	\text{CRPS} = \frac{1}{m T} \sum_{t=1}^{T} \sum_{j=1}^{m} \text{CRPS}_{t, j}.
	$$
	
	\subsection{RMSE-f}
	The RMSE-f(t) quantifies the state forecast error at a prediction horizon of $t$ steps. To compute this metric, we randomly select $P = 1000$ initial conditions $\{x_0^{p}\}_{p=1}^P$ and generate the corresponding forecast ensembles $\{\hat{x}_{t}^{(i), p}\}_{i=1}^{N}$ and $\{x_t^{(i),p}\}_{i=1}^N$ by the surrogate model and the true model, respectively. The ensemble-averaged forecasts are computed as
	$$
	\hat{x}_{t}^{p} = \frac{1}{N} \sum_{i=1}^N \hat{x}_{t}^{(i), p},\  x_{t}^{p} = \frac{1}{N} \sum_{i=1}^N x_{t}^{(i), p}.
	$$
	The RMSE-f(t) is then defined as
	$$
	\text{RMSE-f}(t) = \sqrt{\frac{1}{m P} \sum_{p=1}^{P} \left\| \hat{x}_{t}^{p}-x_{t}^{p}\right\|_2^2}.
	$$
	
	\section{Details for numerical implementations}
	\label{hyperparameter}
	\subsection{Calculation of covariance matrices}
	We introduce a parameterized form of the covariance matrix $Q_{\beta} \in \mathbb{R}^{m\times m}$ in (\ref{transition-model}). For practical implementation, $Q_{\beta}$ is modeled as a diagonal matrix:
	$$
	Q_{\beta} = \begin{bmatrix}
		\text{Softplus}(Q_{\beta,1}) & &\\
		& \ddots & \\
		& & \text{Softplus}(Q_{\beta, m})\\
		\end{bmatrix},
	$$
	where $(Q_{\beta,1}, \cdots, Q_{\beta,m})$ are trainable parameters. The Softplus function is used to ensure the positive definiteness of the covariance matrix. Although a full covariance representation offers greater generality, our numerical experiments show that this diagonal parameterization achieves comparable performance without increased computational complexity.
	
	\subsection{Network architectures}
\begin{figure}[ht!]
		\centering
		\includegraphics[scale=0.17]{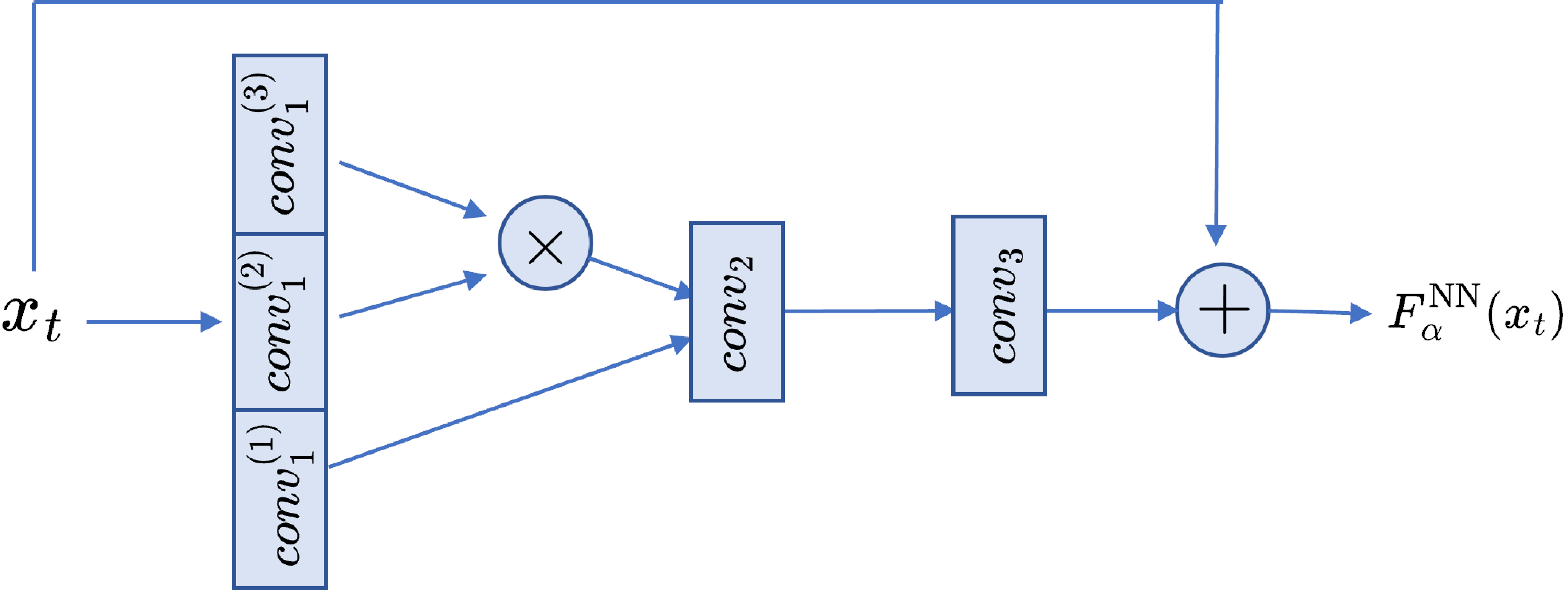}
		\caption{The network architecture of $F_{\alpha}^{\text{NN}}$ for Lorenz 96.}
		\label{96-NN}
	\end{figure}

	We introduce the NN architectures for all methods. For PPO-EnKF, PPO-EnKFc and PPO-PF, the NN architectures consist of two components: an actor and a critic. The actor represents the parameterized surrogate state transition model (\ref{transition-model}), where $Q_{\beta}$ is parameterized as described above. Except for the Lorenz 96 example in Section \ref{Lorenz96}, the deterministic mapping $F_{\alpha}^{\text{NN}}$ is represented using a multi-layer perceptron (MLP). The corresponding hyperparameter settings are provided in Table \ref{tran-setting}. For the Lorenz 96 system, we employ the NN architecture in \cite{EM-EnKS-1}. The specific network design and hyperparameters are illustrated in Figure \ref{96-NN} and Table \ref{96-NN-parameter}. The convolutional layers enables the extraction of local features from high-dimensional inputs, and the bilinear layer is designed to facilitate training when multiplicative interactions are present in the true model. The critic in all experiments is also parameterized using an MLP, with its hyperparameters summarized in Table \ref{tran-setting}. For the EM method, the NN architecture is consistent with the actor across all experiments.
	
	\textbf{Abbreviations:} $N_{\text{layer}}$, the number of fully connected
	layers; $N_{\text{units}}$, the number of hidden units in each layer of the MLP.

	\begin{table}[ht!]
		\centering
		\renewcommand\arraystretch{1.2}
		\begin{tabular}{c|cc|cc|cc|cc}
			\cline{1-9}
			& \multicolumn{2}{c}{\multirow{2}{*}{Actor}} \vline & \multicolumn{6}{c}{Critic}\\
			& \multicolumn{2}{c}{\multirow{2}{*}{}} \vline & \multicolumn{2}{c}{$V_{\phi_1}$} & \multicolumn{2}{c}{$V_{\phi_2}$} & \multicolumn{2}{c}{$V_{\phi_3}$}\\
			\cline{1-9}
			& $N_{\text{layer}}$ & $N_{\text{units}}$ & $N_{\text{layer}}$ & $N_{\text{units}}$ & $N_{\text{layer}}$ & $N_{\text{units}}$ & $N_{\text{layer}}$ & $N_{\text{units}}$\\
			\cline{1-9}
			Section \ref{cm} & 4 & 64 & 2 & 32 & 2 & 32 & 2 & 64\\
			Section \ref{Lorenz63} & 4 & 64 &2 & 32 & 2 & 32 & 2 & 64\\
			Section \ref{Lorenz96} & - & - & 2 & 60 & 2 & 60 & 2 & 120\\
			Section \ref{AC} (PPO-EnKF) & 4 & 100 & 2 & 50 & 2 & 50 & 2 & 100\\
			Section \ref{AC} (PPO-EnKFc) & 4 & 150 & 2 & 75 & 2 & 75 & 2 & 150\\
			\cline{1-9}
		\end{tabular}
		\caption{Hyperparameters of the network structure for the proposed methods.}
		\label{tran-setting}
	\end{table}
	
	\begin{table}[h!]
		\centering
		\renewcommand\arraystretch{1.5}
		\begin{tabular}{cccc}
			\cline{1-4}
			& in-channels & out-channels & kernel-size  \\ \cline{1-4}
			$conv_{1}$ & 1 & 48 & 5  \\
			$conv_{2}$ & 32 & 17 & 5  \\
			$conv_{3}$ & 17 & 1 & 1  \\ \cline{1-4}
		\end{tabular}
		\caption{The hyperparameter settings of $F_{\alpha}^{\text{NN}}$ for Lorenz 96.}
		\label{96-NN-parameter}
	\end{table}
\end{appendices}

\end{document}